%% file: acl_latex.tex
\pdfoutput=1
\documentclass[11pt]{article}
\usepackage{newunicodechar}
\newunicodechar{♫}{\textmusicalnote{}}
\usepackage{amsmath, amsthm}

\usepackage[preprint]{acl}
\usepackage{times}
\usepackage{latexsym}
\usepackage[T1]{fontenc}
\usepackage[utf8]{inputenc}
\usepackage{microtype}
\usepackage{inconsolata}

\usepackage{xspace}
\usepackage{graphicx}
\usepackage{multirow}
\usepackage{amsfonts}

\input{macro.tex}
\title{ECoRAG: Evidentiality-guided Compression for Long Context RAG}

\author{
    Yeonseok Jeong\textsuperscript{1},
    Jinsu Kim\textsuperscript{2}, 
    Dohyeon Lee\textsuperscript{3}, 
    Seung-won Hwang\textsuperscript{3}\thanks{Corresponding Author} \\
    IPAI, Seoul National University\textsuperscript{1}, 
    Korea University\textsuperscript{2}, 
    Seoul National University\textsuperscript{3} \\
    \texttt{\{jys3136, waylight3, seungwonh\}@snu.ac.kr} \\
    \texttt{tonmmy222@korea.ac.kr}
}

\begin{document}
\maketitle

\input{content/0_abstract}
\input{content/1_intro}
\input{content/2_related}
\input{content/3_method}
\input{content/4_experiments}
\input{content/5_analysis}
\input{content/6_conclusion}
\input{content/7_limitation}

\section*{Acknowledgements}
This work was supported by 
the National Research Foundation of Korea(NRF) grant funded by the Korea government(MSIT) (No. RS-2024-00414981), 
Institute of Information \& communications Technology Planning \& Evaluation (IITP) grant funded by the Korea government (MSIT) (No. 2022-0-00077/RS-2022-II220077, AI Technology Development for Commonsense Extraction, Reasoning, and Inference from Heterogeneous Data), and 
Institute of Information \& communications Technology Planning \& Evaluation (IITP) grant funded by the Korea government(MSIT) [NO.RS-2021-II211343, Artificial Intelligence Graduate School Program (Seoul National University)].

\bibliography{anthology,custom}

\newpage
\appendix
\input{content/A_further_analysis}
\input{content/B_details}

\end{document}

%% file: macro.tex
\PassOptionsToPackage{table}{xcolor}
\usepackage{xcolor}
\usepackage{amsmath}
\usepackage{subfigure}
\usepackage{comment}
\usepackage{booktabs} 
\usepackage{caption}
\usepackage{subcaption}
\usepackage{enumitem}
\usepackage{twemojis}
\usepackage{tikz}

\usepackage{tcolorbox}
\usepackage{tabularx}

\usepackage{amssymb}
\usepackage{pifont}

\usepackage{arydshln} 
\usepackage{soul} 

\newcommand{\cmark}{\selectcolormodel{cmy} \textcolor{green}{\ding{51}}}
\newcommand{\xmark}{\selectcolormodel{cmy} \textcolor{red}{\ding{55}}}

\definecolor{mydarkgreen}{HTML}{00CC00} 
\definecolor{mygreen}{HTML}{00CC00}
\definecolor{myblack}{HTML}{000000}
\newcommand{\circled}[1]{%
    \ifnum#1=1
        \tikz[baseline=(char.base)]{
            \node[shape=rectangle,draw=myblack,text=myblack,inner sep=1.5pt,minimum size=10pt] (char) {\textbf{1}};}%
    \else\ifnum#1=2
        \tikz[baseline=(char.base)]{
            \node[shape=rectangle,draw=myblack,text=myblack,inner sep=1.5pt,minimum size=10pt] (char) {\textbf{2}};}%
    \else
        \tikz[baseline=(char.base)]{
            \node[shape=rectangle,draw=gray,text=gray,inner sep=2pt,minimum size=10pt] (char) {\textbf{\romannumeral #1}};}%
    \fi\fi
}

%% file: content/0_abstract.tex
 \begin{abstract}
Large Language Models (LLMs) have shown remarkable performance in Open-Domain Question Answering (ODQA) by leveraging external documents through Retrieval-Augmented Generation (RAG).
To reduce RAG overhead, from longer context, context compression is necessary.
However, prior compression methods do not focus on filtering out non-evidential information, which limit the performance in LLM-based RAG.
We thus propose Evidentiality-guided RAG, or \textbf{ECoRAG} framework.
ECoRAG improves LLM performance by compressing retrieved documents based on evidentiality, ensuring whether answer generation is supported by the correct evidence.
As an additional step, ECoRAG reflects whether the compressed content provides sufficient evidence, and if not, retrieves more until sufficient. 
Experiments show that ECoRAG improves LLM performance on ODQA tasks, outperforming existing compression methods.
Furthermore, ECoRAG is highly cost-efficient, as it not only reduces latency but also minimizes token usage by retaining only the necessary information to generate the correct answer.
Code is available at \href{https://github.com/ldilab/ECoRAG}{https://github.com/ldilab/ECoRAG}.
\end{abstract}

%% file: content/1_intro.tex
\section{Introduction}
\label{sec:introduction}
\input{figure/figure1}

LLMs \cite{openai2023gpt4, touvron2023llama} have excelled in tasks such as ODQA by leveraging external knowledge through RAG \cite{lewis2020retrieval, ram2023context}.
However, RAG inevitably increases context length, which incurs higher computational cost and also hinders generation quality~\cite{liu-etal-2024-lost, hsieh2024ruler, li2024long}.

While adopting existing context compression \cite{li-etal-2023-compressing} may look promising, such a baseline presents two main challenges.
First, LLMs are known to be vulnerable to irrelevant contents that cannot provide evidence for answer generation~\cite{shi2023large, qian2024merge, wu2024faithful}, and existing compression methods \cite{xu2024recomp, jiang-etal-2024-longllmlingua, yoon2024compact} do not effectively filter them out.
As a result, a naive baseline simply prepending retrieved documents, `standard RAG' in Figure \ref{fig:fig1}, outperforms a baseline compressor RECOMP~\cite{xu2024recomp}.
As the number of documents increases, a baseline compressor fails to filter out increasing irrelevant contents, causing performance to decline.

Second, it is challenging to determine the desirable compression ratio for each question.
Failure to do so may lead to compressing too much, which results in losing crucial information, or compressing too little, which produces overly long contexts that degrade generation quality~\cite{liu-etal-2024-lost, hsieh2024ruler, li2024long} and increase computational costs.
Thus, it is necessary to find the desirable compression ratio that enables the LLM to generate the correct answer for each question.

Our distinction is using evidentiality to address both challenges and proposing
\textbf{E}videntiality-guided \textbf{Co}mpression and \textbf{R}etrieval-\textbf{A}ugmented \textbf{G}eneration (\textbf{ECoRAG}) framework: Ours compresses retrieved documents to retain only the information necessary to support the answer. 
To overcome the first challenge, evidentiality \cite{lee2021robustifying, asai2022evidentiality} is used to determine whether each sentence in the retrieved documents supports the correct answer to a question.
It can be quantified for each sentence by measuring how much it contributes to the model to generate the correct answer.
We train the compressor using this as training signals.

To address the second challenge, ECoRAG reflects on compression as a collective, where it contains sufficient evidence.
We begin by forming the smallest possible collective unit of compression and assess whether it is evidential.
If not, it means that it is compressed too much, which we adjust adaptively by collecting more, until it is sufficient.
Through this reflection process, ECoRAG finds the desirable compression ratio that enables the LLM to generate the correct answer with minimal tokens.

By applying these methods, ECoRAG has two advantages when dealing with long contexts as the number of documents increases.
First, ECoRAG improves performance by retaining only the information necessary for generating the correct answer and removing distracting content.
This results in gains on ODQA datasets such as Natural Questions (NQ)~\cite{kwiatkowski2019natural}, TriviaQA (TQA)~\cite{joshi-etal-2017-triviaqa}, WebQuestions (WQ)~\cite{berant2013semantic}.
Second, by compressing the long context to only what is needed, it reduces computational costs.

Our contributions to this work can be summarized as follows: 
(1) Evidentiality-guided Compression: We developed a method that compresses retrieved documents based on evidentiality.
(2) Evidentiality Reflection for Adaptive Compression: Our framework evaluates compressed content for evidentiality and adaptively adjusts the length of compression.
(3) Experiments show that our approach significantly improves retrieval-augmented LLM performance on ODQA datasets.
(4) Our approach is also cost-efficient, as it quickly compresses long context, reducing latency and tokens.

%% file: figure/figure1.tex
\begin{figure}[h]
{
\centering
    \includegraphics[width=0.9\linewidth]{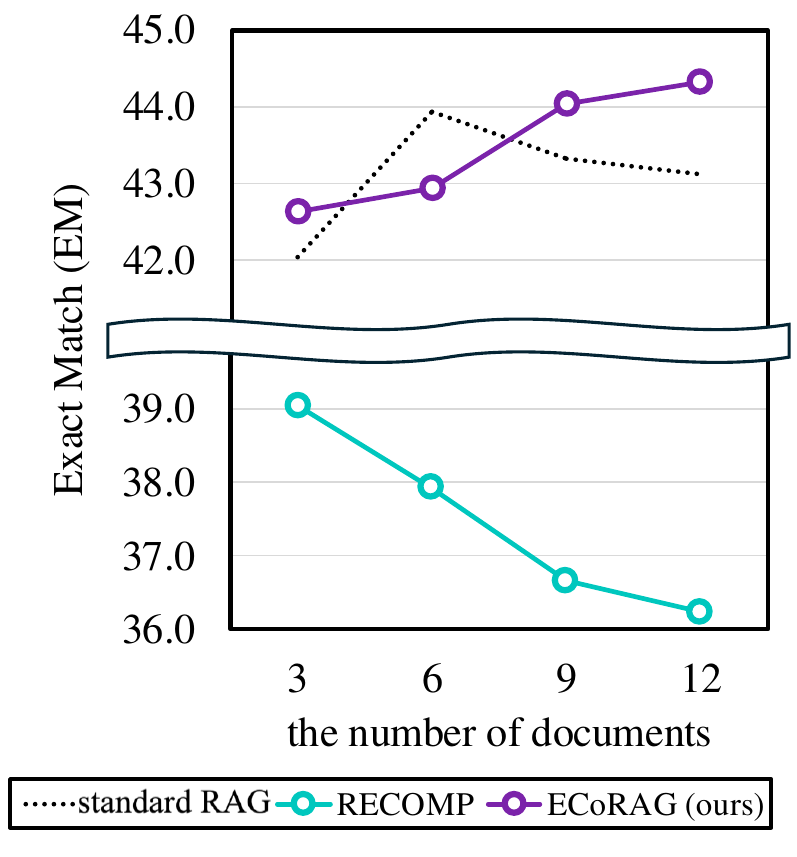}
    \caption{
    Comparison of performance between prepending retrieved documents (standard RAG) \cite{karpukhin-etal-2020-dense}, applying RECOMP \cite{xu2024recomp}, and applying ECoRAG on the Natural Questions \cite{kwiatkowski2019natural} test set.
    Experiments were conducted using Flan-UL2 \cite{tay2023ul}.
    }
    \label{fig:fig1}
}
\end{figure}

%% file: content/2_related.tex
\section{Related Work}
\label{sec:related_work}
\subsection{Evidentiality-guided RAG}
Dense retrievers \cite{karpukhin-etal-2020-dense, izacard2022unsupervised}  focus on lexical answerability, but may mislabel documents as relevant when they lack contextual evidence, leading to the need for evidentiality.
In prior work \cite{lee2021robustifying}, evidentiality refers to whether a document supports generating the correct answer to a question.
Unlike answerability, evidentiality is more challenging to mine directly as it reflects the contextual relationship between a question and a document. 
To measure evidentiality, previous work checks whether the removal of the document is critical for answering the question \cite{asai2022evidentiality}, utilizes attention scores \cite{niu-etal-2020-self}, or considers the change in confidence scores \cite{song2024evidentiality}.
Our work introduces evidentiality in LLMs, enhancing RAG by prioritizing contextually rich documents for generating correct answers.

\subsection{Prompt Compression}
Numerous studies \cite{mu2024learning, li-etal-2023-compressing, kim2024sure} have focused on prompt compression to address both cost and performance challenges, as shown in prior research \cite{shi2023large, liu-etal-2024-lost, hsieh2024ruler}.
RECOMP \cite{xu2024recomp} provides both extractive and generative summaries of documents, considering whether the summaries helped answer the given question. 
LLMLingua \cite{jiang-etal-2023-llmlingua} uses conditional probabilities of LLMs to guide fine-grained prompt compression.
Building on this, LongLLMLingua \cite{jiang-etal-2024-longllmlingua} compresses prompts in long context scenarios by using a question-aware coarse-to-fine compression and document reordering mechanism.
Similarly, CompAct \cite{yoon2024compact} employs an adaptive compression strategy to iteratively compress documents while retaining key information relevant to the query.
However, existing methods struggle to compress long context, which prevents them from fully utilizing the retrieval results. 

\subsection{Retrieval Evaluation for RAG}
LLMs may evaluate the quality of retrieved results for enhancing RAG,
as seen in \citet{madaan2024self}, where models iteratively improve their responses; this concept has been applied to RAG.
Self-RAG \cite{asai2024selfrag} trains LLM to evaluate retrieved documents and its output by predicting reflection tokens that assess the need for retrieval and the quality of the generated text.
\citet{labruna2024retrieve} dynamically determines whether to retrieve additional context when needed by using a trained reader LLM.
CRAG \cite{yan2024corrective} employs a retrieval evaluator to assess document relevance and triggers corrective actions to refine retrieved information, by using lexical overlap between questions and documents.
In our ECoRAG framework, we evaluate whether the evidence is sufficient to generate the correct answer by leveraging evidentiality as defined by the LLM.

%% file: content/3_method.tex
\section{Proposed Method}
\input{figure/main_figure}

In this section, we describe how ECoRAG adaptively adjusts the compression length to ensure that the LLM generates the correct answer.
To achieve this, we focus on:
(1) compressing retrieved documents by sorting them based on evidentiality (Section \ref{sec:3_1}), and 
(2) evaluating whether the compressed documents is sufficiently evidential, and if not, adaptively incorporating more information (Section \ref{sec:3_2}), and Figure \ref{fig:main_figure} provides an overview.

\subsection{Evidentiality-guided Compressor}
\label{sec:3_1}
This section explains how retrieved documents are compressed while preserving the evidence that enables the LLM to generate the correct answer.
We decompose documents into sentences inspired by \citet{xu2024recomp} and compress them guided by evidentiality.
To retain the necessary content and remove irrelevant parts during the compression process, we first extract evidential sentences from the retrieved documents (Section \ref{sec:3_1_1}) and then use them to train the compressor (Section \ref{sec:3_1_2}).

\subsubsection{Definition of Evidentiality}
\label{sec:3_1_1}
We define the evidentiality of a sentence based on its contribution to generating the correct answer while penalizing distractors that interfere with this process.
The degree of evidentiality is categorized hierarchically based on two conditions.
We find sentences that enable the LLM to generate the correct answer.
If a sentence does not, we then check if it interferes with other evidence.

First, when assessing whether each sentence helps generate the correct answer, it is important to consider that the LLM contains parametric knowledge \cite{wang2020language,yu2023generate,luo2023augmented}. 
Prior work \cite{lee2021robustifying, asai2022evidentiality} has focused on whether the language model could contribute to generating the correct answer using given document.
However, it is challenging to distinguish whether the correct answer was generated using the document or parametric knowledge, especially in larger models.
If the correct answer was generated solely using parametric knowledge, regardless of the given document, it is unclear to determine whether the document serves as key evidence.
Therefore, we propose the following first condition:
\circled{1} Without the sentence the LLM cannot generate the correct answer alone, but with the sentence it can.

Second, it is also crucial for the compressor to filter out distractors that hinder the evidence from generating the correct answer. 
While robustness to distractors can be improved through fine-tuning \cite{liu-etal-2024-lost}, training LLMs often requires substantial costs for training and closed LLMs often impossible to train.
If the compressor can remove distractors, it can be applied to any LLM without requiring additional training.
To identify distractors, we introduce a second condition for sentences that do not satisfy \circled{1}: 
\circled{2} The sentence does not interfere with the evidence defined in \circled{1} in generating the correct answer.

\input{figure/evidentiality_mining}

Based on the aforementioned conditions, we hierarchically define evidentiality as depicted in Figure \ref{fig:evidentiality_mining}.
Sentences satisfying condition \circled{1} are labeled as \textbf{strong evidence}. 
Sentences failing to meet condition \circled{1} are further classified based on condition \circled{2}: those satisfying condition \circled{2} are labeled as \textbf{weak evidence}, while those that do not are classified as \textbf{distractor}.
Following these conditions, we use an LLM to label sentences in retrieved documents for each question in the training data.

\subsubsection{Learning Objective for Compressor}
\label{sec:3_1_2}
Given labeled sentences $\mathcal{D}=\{d_1, d_2, \cdots, d_{|\mathcal{D}|}\}$, for a question $q$, we train our compressor based on dual encoders \cite{izacard2022unsupervised} to differentiate between strong and weak evidence, as well as distractor.
Using dual encoders, $E_Q$ for questions and $E_D$ for sentences, we calculate the similarity score between $q$ and sentences in $\mathcal{D}$ (i.e., $sim(q, d_i) = E_Q(q) \cdot E_D(d_i)$).
Sentences are categorized into strong ($d^*$) or weak ($d^+$) evidence, and distractor ($d^-$) based on our hierarchical definition.
We define similarity scores as $s^* = sim(q, d^*)$, $s^+ = sim(q, d^+)$, and $s^- = sim(q, d^-)$.
The similarity scores are utilized to train two inequalities:
\begin{align}
(s^+ > s^-) , \ \ \  (s^* > s^+, s^-)
\end{align}
These inequalities ensure that strong evidence is ranked above weak evidence, which in turn is ranked above distractor, guiding the training of our compressor.

The weak evidentiality loss $\mathcal{L}_{we}$  uses the InfoNCE loss to distinguish weak evidence $d^+$ from distractor $d^-$.
The loss function is formulated as:
\begin{equation}
    \mathcal{L}_{we} = -\log \frac{\exp(s^+/\tau)}{\exp(s^+/\tau) + \sum\limits_{d_j^- \in D^-} \exp(s_j^- / \tau)}
\end{equation}
Here, $s_j^- = sim(q, d_j^-)$ represents the similarity score for each distractor in the set $D^-$, and $\tau$ is a temperature parameter.

The strong evidentiality loss $\mathcal{L}_{se}$ also utilizes the InfoNCE loss to prioritize strong evidence $d^*$.
The loss function is formulated as:
\begin{equation}
    \mathcal{L}_{se} = -\log \frac{\exp(s^*/\tau)}{\exp(s^*/\tau) + \sum\limits_{d_j^{\pm} \in D^-\cup D^+} \exp(s_j^{\pm}/\tau)}
\end{equation}
Here, $s_j^{\pm} = sim(q, d_j^-)$ is the similarity score for each sentence in the combined sets of distractors $D^-$ and weak evidences $D^+$.

The final loss $\mathcal{L}$ is defined as the sum of the strong and weak evidentiality losses: 
\begin{equation}
    \mathcal{L} = \mathcal{L}_{se} + \mathcal{L}_{we}
\end{equation}
Our compressor is trained using this loss $\mathcal{L}$, and ranks sentences $d'_1, d'_2, \dots, d'_{|\mathcal{D}|}$ by evidentiality, selecting high-scoring ones for compression.
The number of sorted evidence required can vary depending on the difficulty of each question.
However, providing too little evidence may omit important information, while too much increases computational costs for each question.
Thus, balanced compression ratio is necessary for each question to address both issues.

\subsection{Evidentiality Reflection for Adaptive Compression}
\label{sec:3_2}
Once a collective of evidential sentences is formed, we need to determine whether the compression ratio is appropriate.
To achieve this, we reflect on the evidentiality of compressed documents using a language model (Section \ref{sec:3_2_1}).
Then, if compressed too much, we adaptively adjust the compression ratio by collecting more (Section \ref{sec:3_2_2}).

\subsubsection{Training Evidentiality Evaluator}
\label{sec:3_2_1}
We develop an effective \textbf{evidentiality evaluator} $\mathcal{M}_{eval}$ that assesses whether the compressed documents are strong evidence enough to generate the correct answer. 
In prior work, CompAct \cite{yoon2024compact} trained the evaluator by prompting GPT-4o \cite{openai2023gpt4} to determine if the evidence is sufficient to answer the question.
However, this approach can introduce bias \cite{chiang-lee-2023-large} when GPT-4o evaluates through prompting, leading to inaccurate supervision.
Accurate supervision requires verifying if the document actually enables the reader LLM to generate the correct answer.
To achieve this, we reuse our evidentiality labels obtained from the LLM in Section \ref{sec:3_1_1} and distill them from our reader LLM into smaller model, Flan-T5-large \cite{chung2022scaling}, to build the evaluator. 
Comparison between CompAct and our evaluator is discussed in Section \ref{sec:5_2}.

We train $\mathcal{M}_{eval}$ using our evidentiality labeled dataset $(d^*, d^+, d^-)$ to determine if compressed documents are sufficient for correct answer generation.
The evaluator is trained to classify whether the given compressed documents is strong evidence.
To facilitate this, we add 2 special tokens $t\in [\text{<EVI>}, \text{<NOT>}]$ and train $\mathcal{M}_{eval}$ to generate `<EVI>' for strong evidence $d^*$, and `<NOT>' for other sentences $d^+, d^-$. 
Subsequently, next-token prediction loss $\mathcal{L}_{eval}$ is used for this training stage to predict whether compressed documents are strong evidence.
\begin{equation}
    \mathcal{L}_{eval}=- log\ p_{\mathcal{M}_{eval}}(t|q,d)
\end{equation}

\subsubsection{Adaptive Compression}
\label{sec:3_2_2}
In adaptive compression, the compression ratio is adaptively adjusted by our evaluator, which reflects on whether the current compression is evidential, as described in Figure \ref{fig:main_figure}.
Initially, our evaluator assesses the evidentiality of compressed documents $C$ containing only the first evidence, $d'_1$, from our ordered evidences $d'_1, d'_2, \dots, d'_{|\mathcal{D}|}$.
If the evaluator determines that $C$ is evidential, it becomes the final compression provided to LLM.
If $C$ is not evidential, we add the next piece of evidence $d'_2$ to $d'_1$ to build new compressed documents.
If the k-th iteration fails, $d'_{k+1}$ is added to the previously compressed documents.
This process is repeated until the desirable compression is found, with a token limit set to avoid infinite loop.
Since retrieved documents do not always include gold evidence for all queries, a token limit is necessary to prevent infinite loops from continuously adding evidence.
The final compression is then used as input for the LLM, which generates the final answer.

Although iterative adjustment can increase latency compared to using raw documents, ECoRAG reduces it efficiently.
Prior work \cite{yoon2024compact}, each iteration required LLM (7B) to generate a new compression by using the previous compression and the next piece of evidence. 
Thus, with each iteration, LLM reads different contents and generates compression of multiple tokens, increasing latency time.
However, ECoRAG reduces redundancy by ordering evidence just once and adding it iteratively.
Moreover, our framework utilized a lightweight evaluator (0.77B) that adjusts compression length by generating just a single special token, resulting in rapid compression speed; the actual results are shown in Section \ref{sec:5_4}.

%% file: figure/main_figure.tex
\begin{figure*}[h]
{
\centering
    \includegraphics[width=1.0\linewidth]{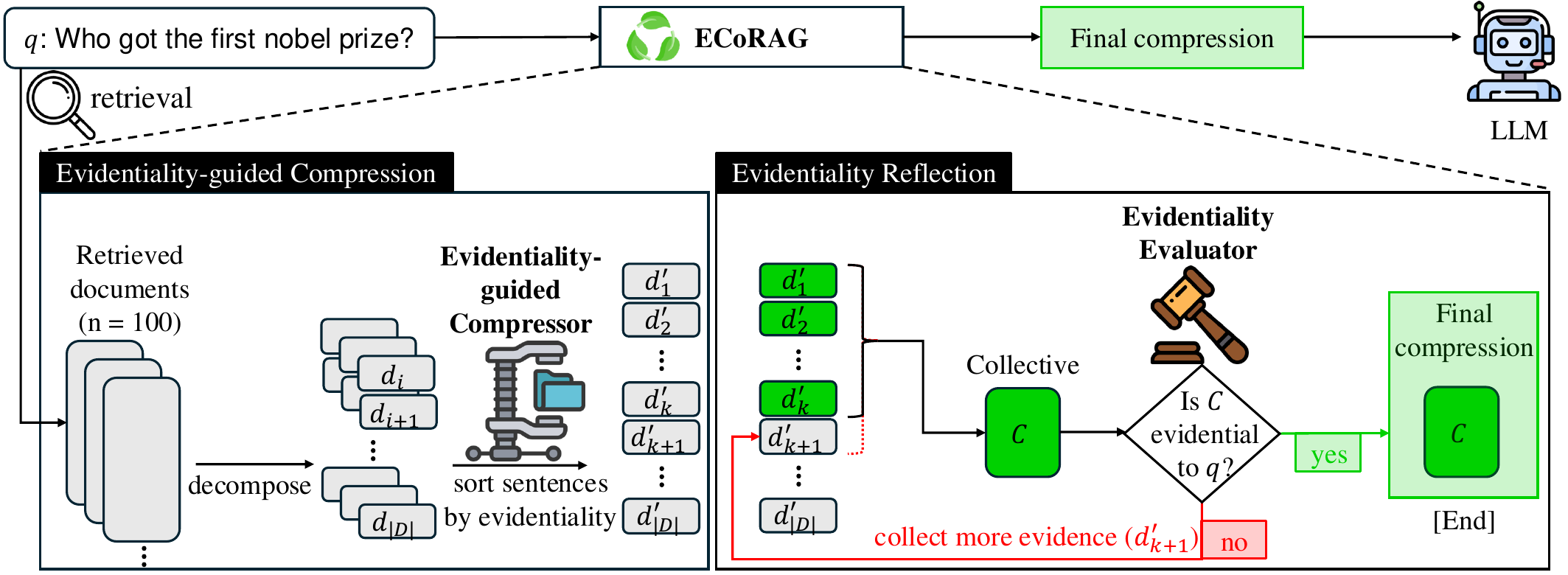}
    \caption{
    This figure illustrates the overall framework of ECoRAG. 
    First, the evidentiality-guided compressor compresses the retrieved documents by sorting decomposed sentences based on evidentiality, producing an ordered set of evidences $d'_1, d'_2, \dots, d'_{|\mathcal{D}|}$.
    Second, evidentiality reflection starts with the top-ranked sentence ($n=1$, i.e., $C=d'_1$), and the evidentiality evaluator determines whether $C$ is evidential.
    If not, more evidence is added iteratively ($n=k\rightarrow n=k+1$) until the evaluator judges $C$ as evidential.
    Once evidential, it is used for final compression (green line); otherwise, additional evidence is collected (red line).}
    \label{fig:main_figure}
}
\end{figure*}

%% file: figure/evidentiality_mining.tex
\begin{figure}[h]
{
\centering
    \includegraphics[width=1.0\linewidth]{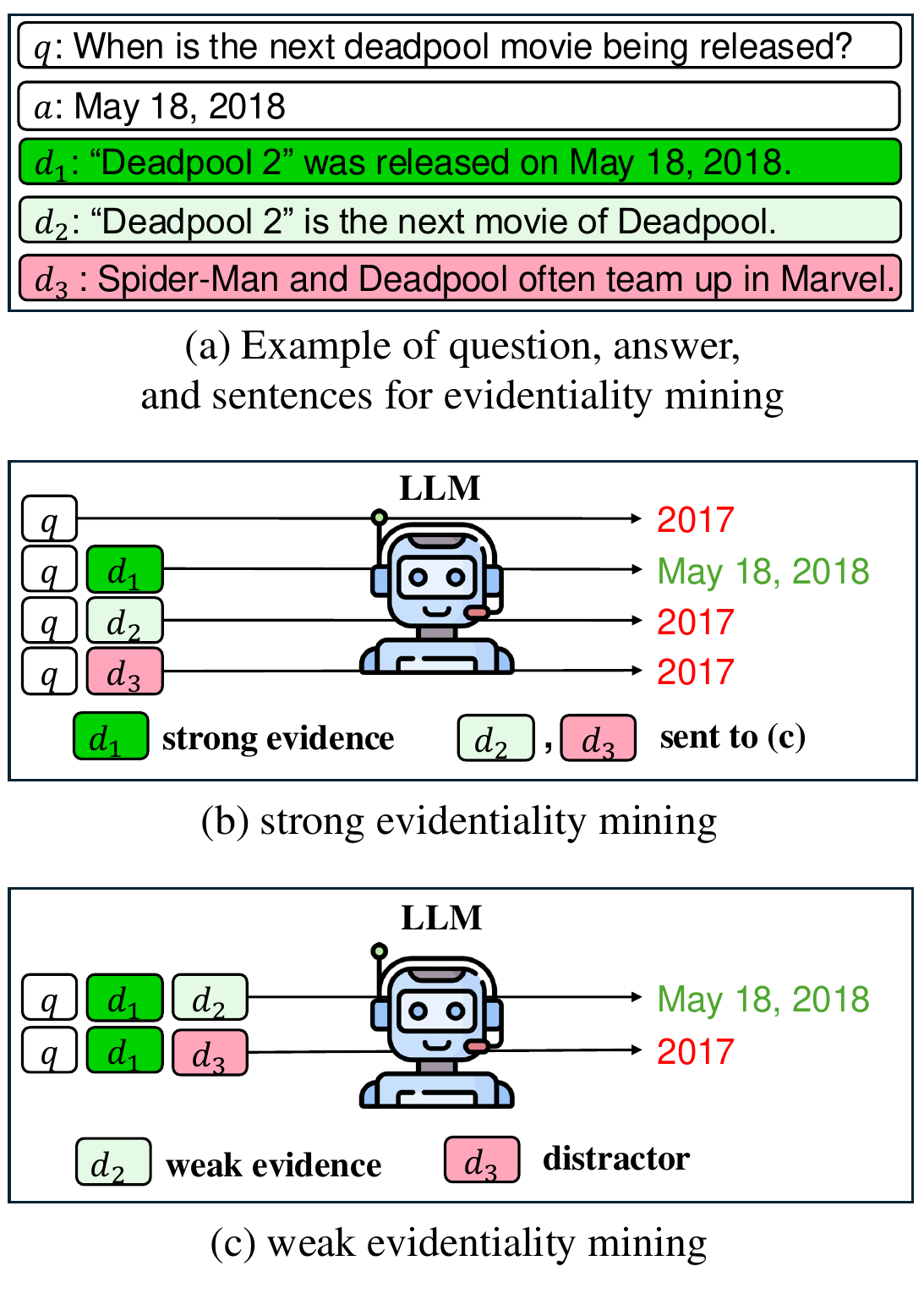}
    \caption{
    This figure illustrates the evidentiality mining strategy of  ECoRAG.
    }
    \label{fig:evidentiality_mining}
}
\end{figure}

%% file: content/4_experiments.tex
\section{Experiments}
\subsection{Experimental Settings}
\input{table/gpt_results}

\paragraph{Datasets}
We evaluate our framework through NQ \cite{kwiatkowski2019natural}, TQA \cite{joshi-etal-2017-triviaqa}, and WQ \cite{berant2013semantic}, which are ODQA datasets. 
We use the 100 documents retrieved from DPR \cite{karpukhin-etal-2020-dense}\footnote{Since enhancing the retriever is beyond the scope of this study, we conduct our experiments under the assumption that the retrieved documents are already provided.}.

\paragraph{Models} 
We initialize our evidentiality compressor from Contriever~\cite{izacard2022unsupervised} and use it to compare its performance with RECOMP~\cite{xu2024recomp}.
For evidentiality evaluator, we utilize Flan-T5-large \cite{chung2022scaling}, because previous RAG and document-assessment work~\cite{han2023monotonic, yan2024corrective} have successfully employed it.
Detailed justification for this choice can be found in Section \ref{sec:b_2}.
For the reader model, we use GPT-4o-mini~\cite{openai2023gpt4}, as it supports a context length of 128K tokens, sufficient to process all 100 retrieved documents.

\paragraph{Evaluation Metrics} 
We report results on the test sets of NQ, TQA, and WQ using EM and word-level F1-score to assess the question-answering task performance. 
We also report the average number of input tokens given to the reader LLM to evaluate the efficiency of our compression step.

\paragraph{Baseline} We report two types of baselines.

\textbf{\textit{RAG without compression}}: As a baseline, we report the results using only the question and raw retrieved documents.
The `closed-book' setting, where no retrieval is used, shows that the model relies solely on its internal knowledge.
In the `standard RAG' setting, we simply concatenate the top 100 retrieved documents without any compression for evaluation.\footnote{We also evaluated the effect of reducing the number of retrieved documents ($k=5,10,20$) for both DPR and Contriever in Table \ref{tab:various_doc_size}; results are explained in Section \ref{sec:a_9}.}
This is the approach used in conventional RAG without compression.

\textbf{\textit{RAG with 100 compressed documents}}:
We also reproduce several retrieval augmentation methods for comparison.
To better understand the effect of different compression methods, we evaluated several baselines including LLMLingua~\cite{jiang-etal-2023-llmlingua}, LLMLingua-2~\cite{pan-etal-2024-llmlingua}, LongLLMLingua~\cite{jiang-etal-2024-longllmlingua}, CompAct~\cite{yoon2024compact}, and RECOMP which offers both extractive and abstractive variants.

In addition to our compression and non-compression baselines, we include BGE-M3~\cite{chen2024bge} and BGE-reranker~\cite{xiao2024c} under equal token budgets.
However, since these are not compression methods, their comparison results are addressed in Section \ref{sec:a_4}.

\subsection{Results}
\label{sec:4_2}
In this section, we report the results of our model and compare them with both compression-based and non-compression baselines for ODQA in Table \ref{tab:gpt_result}.
Accuracy, such as EM and F1-score, is a more important metric than token reduction for evaluating compression quality because simply reducing tokens without preserving necessary information is meaningless.
A method is more efficient if it reduces more tokens while maintaining higher accuracy than another.

In terms of accuracy, ECoRAG outperforms all baselines, including standard RAG, where the LLM reads all retrieved information.
In the long context setting, retrieving many documents often brings in those with low relevance scores, introducing noise. 
However, previous compression methods fail to filter out this noise, leading to performance degradation compared to uncompressed approaches.
Notably, ECoRAG surpasses all these methods, even with fewer tokens than some of them. 
The strength of ECoRAG lies in compressing only the necessary content, focusing solely on the information essential for generating the correct answer.
As a result, ECoRAG outperforms the strongest compression baseline in NQ (+0.77\%p), TQA (+1.38\%p), and WQ (+0.40\%p) in EM.
As further detailed in Section~\ref{sec:a_10}, ECoRAG maintains this advantage even on much longer retrieved documents, confirming its robustness in another long context setting~\cite{bai2024longbench}.

From a token efficiency perspective, ECoRAG uses more tokens than RECOMP (abstractive) and CompAct but still outperforms them, while compressing with fewer tokens than other methods.
According to \citet{xu2024recomp}, abstractive RECOMP performs well in the 5-document setting but struggles in long contexts due to input size limitations.
CompAct suffers from inaccurate compression evaluation, failing to retain essential information, which lowers performance.
In contrast, ECoRAG can handle long context and retain only the necessary content to generate the correct answer, which results in superior performance across different datasets.
Excluding the two compressors that fail to preserve necessary information, ECoRAG achieves higher accuracy with fewer tokens than other methods, demonstrating its token efficiency.

%% file: table/gpt_results.tex
\begin{table*}[]
    \centering
    \scalebox{0.77}{
    \begin{tabular}{lcccccccccc}
    \noalign{\hrule height 1pt}
    \multirow{2}{*}{\textbf{Methods}} & \multicolumn{3}{c}{\textbf{NQ}} & \multicolumn{3}{c}{\textbf{TQA}} & \multicolumn{3}{c}{\textbf{WQ}} \\
                             &\textbf{\#tokens $\downarrow$}  & \textbf{EM}     & \textbf{F1} & \textbf{\#tokens $\downarrow$}   & \textbf{EM}   & \textbf{F1} & \textbf{\#tokens $\downarrow$}   & \textbf{EM}   & \textbf{F1} \\ \hline
    \textbf{\textit{RAG without compression}} & \\
    ~~ closed-book       & 0        & 31.88  &  44.10  &         0 &     64.78 & 73.10 & 0 & 24.51 & 42.73 \\
    ~~ standard RAG (100 documents)    &    13905 & 36.09  &  \textbf{50.18}  &     14167 & 56.21 & 64.22 & 13731 & 21.11 & 38.72 \\ \hline
    \textbf{\textit{RAG with 100 documents compressed}} & \\
    ~~ LLMLingua~\cite{jiang-etal-2023-llmlingua}              &   635 &   26.84 &   38.30 &       630 & 50.81 & 57.91 & 641 & 22.98 & 39.77 \\
    ~~ LLMLingua-2~\cite{pan-etal-2024-llmlingua}           &     1315 &   30.11 &  42.52  &      1324 & 53.19 & 60.46 & 1113 & 23.52 & 40.61 \\
    ~~ LongLLMLingua~\cite{jiang-etal-2024-longllmlingua}           &    1370 &   32.96 &   45.32 &      1402 & 55.75 & 63.75 & 1355 & 21.51 & 39.13 \\
    ~~ RECOMP (extractive)~\cite{xu2024recomp}   &      662 &   32.85 &   44.54 &       672 & 51.66  & 59.08 & 658 & 19.54 & 36.83 \\
    ~~ RECOMP (abstractive)~\cite{xu2024recomp}   &      \textbf{14} &  27.59 &   39.19 &  \textbf{26} & 39.95 & 46.68 & \textbf{19} & 20.47 & 36.90 \\
    ~~ CompAct~\cite{yoon2024compact}     &         106 &   35.71 &   47.14 &         96 &   63.96   &   73.87 & 75 & 29.77 & 44.25 \\
    ~~ ECoRAG (ours)             &         632 & \textbf{36.48}  &  49.81  &          441 &   \textbf{65.34}   &   \textbf{75.37}  & 560 & \textbf{30.17} & \textbf{46.13} \\
    \noalign{\hrule height 1pt}
    \end{tabular}
    }
    \caption{Compression methods performance comparison on  NQ, TQA, and WQ. The table shows the results using GPT-4o-mini as the reader model, given 100 retrieved documents~\cite{karpukhin-etal-2020-dense}. It reports the number of tokens after compression, along with EM and F1-score, illustrating the impact of different compression methods on model performance.}
    \label{tab:gpt_result}
\end{table*}

%% file: content/5_analysis.tex
\section{Analysis}
\label{sec:analysis}
In addition to the main results, we verified the effectiveness of our framework by addressing the following research questions:
\begin{itemize}
\item \textbf{RQ1}: Does our compressor effectively capture human-annotated evidence?
\item \textbf{RQ2}: How accurately does our evaluator predict evidentiality?
\item \textbf{RQ3}: What is the impact of each component in ECoRAG?
\item \textbf{RQ4}: Is ECoRAG efficient compression?
\end{itemize}

\subsection{RQ1: Alignment with Human-annotated Evidentiality}
\label{sec:5_1}
\input{table/compressor_hotpotqa}
In this section, we assess whether our compressor can effectively sort sentences by evidentiality for next step.
Although our compressor improves LLM performance by learning LLM-defined evidentiality, it is essential to verify whether it effectively captures ground-truth evidence.
Thus, we conducted experiments using HotpotQA \cite{yang-etal-2018-hotpotqa}, which provides human-annotated evidence.
We compared how well prior methods and our compressor assign higher scores to ground-truth evidence.
For evaluation, we use Normalized Discounted Cumulative Gain (NDCG) as a metric to evaluate how effectively evidentiality-focused methods, including ours, rank evidence higher.

As shown in Table \ref{tab:human-align-hotpot}, ECoRAG achieved the highest performance, demonstrating strong alignment with human-annotated evidentiality.
The `Answerability' baseline trains the compressor by treating passages containing the correct answer as positive and those without as negative.
The `Leave-One-Out' \cite{asai2022evidentiality} considers a passage as positive if removing it prevents the model from generating the correct answer, and negative if the model still succeeds. 
ECoRAG outperforms prior evidentiality baselines, achieving improvements in NDCG@1 (+4.86\%p) and NDCG@10 (+1.12\%p)
This result indicates that our compressor effectively captures evidence and aligns well with human annotations.
Thus, our compressor provides well-sorted evidences to our evaluator, then we need to verify the evaluator, the other component of ECoRAG.

\subsection{RQ2: Evaluator Performance on Evidentiality Prediction}
\label{sec:5_2}
\input{figure/evaluator_metrics}

We also need to verify the evidentiality evaluator to accurately evaluate whether the compressed documents enable the LLM to generate the correct answer.
To assess its accuracy, we conducted experiments on the TQA test set.
For each question, we define ground-truth labels for retrieved documents as either <EVI>, which lead to generating the correct answer as in Section \ref{sec:3_2_1}, or <NOT>.
We then measured how well our evaluator and other evaluators predicted these labels using accuracy, precision, recall, and F1-score.
The results are shown in Figure~\ref{fig:evaluator_metrics}.

Across all metrics, our evidentiality evaluator effectively predicts evidentiality, even though it has significantly fewer parameters than other evaluators.
It outperforms the CompAct evaluator (7B)~\cite{yoon2024compact} by +13.96\%p in F1 score.
The CompAct evaluator is based on Mistral-7B~\cite{jiang2023mistral} and trained with supervision from GPT-4o.
As \citet{asai2024selfrag} noted, the reader LLM evaluates whether documents support the correct answer, making it a strong baseline.
We used Flan-UL2 \cite{tay2023ul} (20B) as our reader LLM, as described in Section \ref{sec:prompt}.
Notably, our evidentiality evaluator, despite its much smaller size (770M), closely approximates the performance of Flan-UL2 (-0.08p\%).

\subsection{RQ3: Ablation Study}
\label{sec:5_3}

\input{table/ablation}
In Table \ref{tab:ablation}, we present the results of our ablation study, assessing the impact of each component in our framework by comparing EM across different settings.
We also report R20, checking if the gold answer is in the top 20 sentences.

For \textbf{\textit{Compressor}}, we compare (A) ECoRAG with two inferior compressors, (B) and (C). 
In (B), the compressor uses a pretrained Contriever checkpoint without additional training, while in (C), it is trained with answerability labels.
As shown, our compressor trained with evidentiality labels outperforms both alternatives.
Comparing (A) and (C) shows that evidentiality labels increase EM (+1.02\%p, +0.53\%p) while maintaining R20 at a comparable level.
Since R20 measures lexical overlap, (C), trained with answerability, performs similarly to or better than (A).
The results demonstrate the superiority of our evidentiality labels over answerability labels, as they prioritize contextually rich information. 

For \textbf{\textit{Evaluator}}, we consider a no-evaluator setting (D), where the initial compression from the compressor is used without evaluating its evidentiality.
The EM gap between (A) and (D) (+0.77\%p, +1.80\%p) highlights the impact of the evidentiality evaluator.
These results highlight the importance of adaptively adjusting the amount of evidence through evidentiality evaluation.

\subsection{RQ4: Total Latency}
\label{sec:5_4}
\input{table/speed_test}
ECoRAG is cost-efficient not only because it reduces the number of tokens but also because it decreases total latency in the RAG process.
In RAG without compression, computational costs increase as more documents are retrieved.
By applying compression and retaining only the necessary information, ECoRAG reduces total processing time.

Table~\ref{tab:speed_test} presents the total latency\footnote{Since GPT-4o-mini does not provide latency measurements, we conducted the latency experiments using Flan-UL2.}, including both compression and inference time, to show the efficiency of our approach.
For long context, the LLM-based abstractive compressor CompAct took longer than the `standard RAG' setting, whereas the extractive compressors RECOMP and ECoRAG were faster.
ECoRAG uses the lightweight evaluator that generates only a single token per iteration, stopping the reflection process once the compressed document is evidential or the token limit is reached, thereby preventing excessive compression time.
While ECoRAG had similar speed to RECOMP, it achieved better performance by retaining only the information necessary to generate the correct answer, as described in Table \ref{tab:gpt_result}.
Thus, ECoRAG is effective in handling long contexts in terms of both performance and efficiency.

ECoRAG is a two-step design that achieves both speed and performance.
Single-step aggregation with LLMs, as demonstrated by CompAct in Table~\ref{tab:gpt_result}, struggles with length dependency for listwise evaluation due to the ``lost-in-the-middle'' issue~\cite{liu-etal-2024-lost}.
In contrast, ECoRAG separates the process by first assessing sentences individually with an extractive compressor and then evaluating them collectively. 
This separation overcomes challenges in handling long contexts and improves compression effectiveness. 
Our lightweight components ensure efficiency while achieving effective compression.

%% file: table/compressor_hotpotqa.tex
\begin{table}[t]
  \centering
  \resizebox{1.0\linewidth}{!}{
  \begin{tabular}{lcc}
  \noalign{\hrule height 1pt}
  \textbf{Methods}                             & \textbf{NDCG@1} & \textbf{NDCG@10} \\
  \hline
  Answerability (baseline)           & 67.82 & 79.20 \\
  Leave-One-Out \cite{asai2022evidentiality}                      & 70.67 & 80.80 \\
  ECoRAG (ours)     & \textbf{75.53} & \textbf{81.92} \\
  \noalign{\hrule height 1pt}
  \end{tabular}
  }
  \caption{Comparison of NDCG@1 and NDCG@10 on HotpotQA dataset using different training signals}
  \label{tab:human-align-hotpot}
\end{table}

%% file: figure/evaluator_metrics.tex
\begin{figure}[h]
{
\centering
    \includegraphics[width=0.8\linewidth]{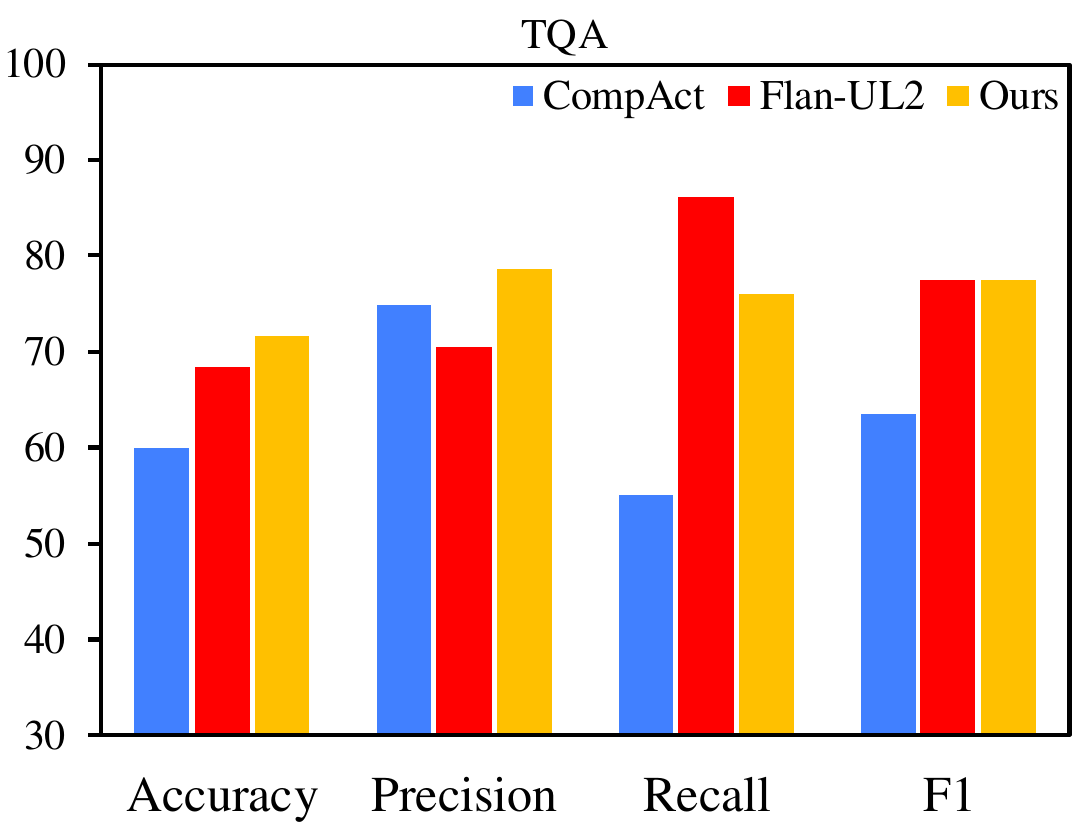}
    \caption{
    Evidentiality evaluation metrics using different evaluator, including ours, measured on the TQA.
    }
    \label{fig:evaluator_metrics}
}
\end{figure}

%% file: table/ablation.tex
\begin{table}[t]
\newcommand{\deepgrey}[1]{\textcolor[RGB]{128,128,128}{\textbf{#1}}}
  \centering
  \resizebox{0.98\linewidth}{!}{
  \begin{tabular}{lcccccc}
  \noalign{\hrule height 1pt}
                                       & \multicolumn{2}{c}{\textbf{NQ}} & \multicolumn{2}{c}{\textbf{TQA}}  \\
                                       & \textbf{EM} & \textbf{R20} & \textbf{EM} & \textbf{R20} \\
  \hline
   (A) ECoRAG (ours)                          & \textbf{36.48} & \textbf{75.18} & \textbf{65.43} & 80.38  \\
   \hline
   \textbf{\textit{Compressor}} \\
   \text
   ~~ (B) w/o answerability         & 31.25 & 49.53 & 63.86 & 70.84 \\
   ~~ (C) w/o evidentiality  & 35.46 & 74.93 & 64.90 & \textbf{80.59} \\
   \hline
   \textbf{\textit{Adaptive Compression}}                            \\
   ~~ (D) w/o evaluator    & 35.71 & - & 63.63 & - \\
  \noalign{\hrule height 1pt}
  \end{tabular}  
  }
  \caption{Ablation study of ECoRAG, showing the impact of compressor and adaptive compression methods.}
  \label{tab:ablation}
\end{table}

%% file: table/speed_test.tex
\begin{table}[t]
    \centering
    \scalebox{0.65}{
    \begin{tabular}{lcccc}
        \noalign{\hrule height 1pt}
        \textbf{Methods} 
        & \begin{tabular}[c]{@{}c@{}}\textbf{Compression} \\ \textbf{Time}\end{tabular} 
        & \begin{tabular}[c]{@{}c@{}}\textbf{Inference} \\ \textbf{Time}\end{tabular} 
        & \begin{tabular}[c]{@{}c@{}}\textbf{Total} \\ \textbf{Time}\end{tabular} 
        & \begin{tabular}[c]{@{}c@{}}\textbf{Throughput} \\ \textbf{(example/sec)}\end{tabular} \\
        \hline
        closed-book & - & 3.79h & 3.79h & 0.26  \\
        standard RAG & - & 12.28h & 12.28h & 0.08  \\
        RECOMP & 0.27h & 4.08h & 4.35h & 0.23  \\
        CompAct & 10.10h & 4.83h & 14.94h & 0.07  \\
        ECoRAG (ours) & 0.73h & 4.23h & 4.96h & 0.20  \\
        \noalign{\hrule height 1pt}
    \end{tabular}}
    \caption{Inference time and compression time for NQ test.}
    \label{tab:speed_test}
\end{table}

%% file: content/6_conclusion.tex
\section{Conclusion}
ECoRAG is a framework designed to compress long context by focusing on evidentiality in LLMs, defined as whether information supports generating the correct answer. 
Evidentiality-guided compression effectively filters out irrelevant content and retains necessary evidence.
Through adaptive compression, ECoRAG determines the optimal compression length for each question, ensuring efficient use of context.
As a result, ECoRAG demonstrates both superior performance and efficiency in handling long context, outperforming other compression methods.

%% file: content/7_limitation.tex
\section{Limitation}
Evidentiality provides an effective indicator for determining whether information is necessary for an LLM to generate the correct answer.
However, mining evidentiality labels is computationally expensive, leading to increased costs. 
Since multiple inferences are required for each question, it results in significant time consumption.
Nevertheless, as more time is spent, more evidentiality labels can be obtained, which can contribute to the training of the compressor.
Evidentiality labels can also be reused to train the evidentiality evaluator, optimizing resource usage. 
Once the compressor is fully trained and applied, the LLM inference process becomes faster.

Building upon this efficiency improvement, the application of this system can be extended beyond ODQA to address broader real-world scenarios.
Extending it to tasks like summarization may be necessary due to context length limits when processing full content with LLMs.
Selecting and summarizing only the most important parts can improve performance~\cite{saxena2024select, jeong2025agent}, requiring evidentiality to be redefined based on summarization metrics.
Investigating such adaptations is a potential direction for future work.

%% file: content/A_further_analysis.tex
\clearpage
\section*{\centering Appendices}
\section{Further Analysis}
\subsection{Comparative Analysis of Compression Methods}
\input{table/contribution}
In this section, we will provide a more detailed comparison of our approach with other baselines based on Table \ref{tab:gpt_result}.
Table \ref{tab:comparison} provides an overview of how each method differs.
Based on this comparison, we discuss how large-scale documents can be compressed efficiently and effectively.

In ODQA, since the model must provide an answer to a given question, the compression process needs to consider the question.
LLMLingua \cite{jiang-etal-2023-llmlingua} and LLMLingua-2 \cite{pan-etal-2024-llmlingua}, which do not consider the question during compression, often include irrelevant information, leading to suboptimal performance.
On the other hand, the methods other than LLMLingua and LLMLingua-2 are question-aware, allowing them to more effectively capture the necessary content, resulting in higher performance compared to question-agnostic methods.

The amount of evidence needed varies for each question, and one solution to address this is adaptive compression, where the compression ratio is adjusted for each question.
By applying this method, only the necessary tokens are used, leading to high performance with fewer tokens. 
As seen in Table \ref{tab:gpt_result}, both CompAct \cite{yoon2024compact} and ECoRAG achieve high performance with a reduced number of tokens.

However, there are two main challenges when dealing with long context.
First, while using numerous retrieval results increases the amount of necessary information available, it also includes documents with lower relevance scores, resulting in considerable noise.
Second, the overall length of the documents is too long, which makes the compression process time-consuming.

To address the first challenge mentioned above, the concept of evidentiality is necessary.
As discussed in Section \ref{sec:3_1_1}, by prioritizing strong evidence for correct answer generation and penalizing distractors, we have been able to create a compressor that is robust against noise.
Consequently, this approach allows ECoRAG to demonstrate the highest performance in large-scale document settings.

To address the second challenge, the compressor must be an extractive compressor that evaluates each content pointwise and extracts only the necessary information.
Language model-based abstractive compressor is hindered by limited context length, which leads to truncation and fails to handle entire large-scale documents.
Moreover, LLM-based abstractive compressor often requires substantial time for inference and may suffer from positional biases \cite{liu-etal-2024-lost}, which can lead to inaccurate assessments of evidentiality. 
However, extractive compressors such as ECoRAG and RECOMP (extractive) \cite{xu2024recomp} are lightweight models that can quickly calculate scores, as seen in Table \ref{tab:speed_test}, and process each document in parallel for each document, thus avoiding positional biases.

Based on these observations, we conclude that ECoRAG, which combines all the characteristics from Table \ref{tab:comparison}, is appropriate for compressing large-scale documents effectively.

\subsection{Evaluator Performance on NQ}
\input{figure/evaluator_metrics_nq}

We conducted same experiments on NQ~\cite{kwiatkowski2019natural}, as described in Section \ref{sec:5_2}, observed similar trends to those in TQA~\cite{joshi-etal-2017-triviaqa}.
As shown in Figure \ref{fig:evaluator_metrics_nq}, our evidentiality evaluator consistently outperforms CompAct and demonstrates comparable results to Flan-UL2, further validating its effectiveness across different datasets.

\subsection{Compression Effectiveness with More Long Context}
\label{sec:1000_doc}
\input{table/1000_doc}
To explore performance of ECoRAG with more documents, we conducted additional experiments using 1000 retrieved documents in Table~\ref{tab:1000_doc}. 
Previous compression work, such as CompAct, focused on up to 30 documents, while our experiments used 100 documents, a common setting in RAG models like FiD~\cite{izacard2020leveraging}. 
To verify whether our method consistently improves performance even with more documents, we tested with 1000 documents. 
Due to limited budget, we used documents already retrieved by a DPR setting that was searched, differing from our top-100 DPR setting. 
We compared ECoRAG with RECOMP, an extractive method with a similar structure, and excluded abstractive compressors such as CompAct due to its too high latency in longer context compression.

With 1000 documents, ECoRAG remained highly effective in compressing and preserving essential information.
The context length became too long for GPT-4o-mini to effectively utilize the information~\cite{hsieh2024ruler}, as shown in Table~\ref{tab:1000_doc}. 
However, our compression effectively reduced the length, maintaining high performance. 
Additionally, ECoRAG outperformed other extractive compressors, demonstrating its superiority in handling extensive document sets.

ECoRAG remains the most effective compressor even for extremely long contexts.
Without compression, excessive context length can degrade performance or exceed the context limit.
In contrast, our retriever-based compressor efficiently compresses extended inputs regardless of length.

\subsection{A Comparative Study with Reranker}
\label{sec:a_4}
\input{table/bge-m3}
ECoRAG fundamentally differs from reranking methods like BGE-M3~\cite{chen2024bge} and RECOMP by adaptively determining the rank and compression ratio needed for each query. 
While reranking models focus on relevance, they lack our ability to iteratively refine compression based on evidentiality. 
To ensure a fair comparison with our approach in terms of token usage, we conducted additional experiments with both BGE-M3\footnote{\href{https://huggingface.co/BAAI/bge-m3}{BAAI/bge-m3}} and BGE-reranker\footnote{\href{https://huggingface.co/BAAI/bge-reranker-large}{BAAI/bge-reranker-large}}~\cite{xiao2024c} by using its reranked top-10 and top-20 sentences.
As shown in Table~\ref{tab:bge}, ECoRAG achieves better performance, demonstrating the importance of selecting the appropriate context over simply increasing or reducing the amount of information.

Unlike other sentence reranking methods, ECoRAG evaluates the initial compression and adaptively adjusts the compression ratio through a reflection process to determine how much information is required. 
This capability moves ECoRAG closer to true compression rather than simple reranking.
Furthermore, our research extends beyond proposing a compressor—it introduces a complete framework. 
While we used Contriever to ensure fair comparisons with RECOMP, our framework is flexible and capable of training models like BGE-M3 and BGE-reranker to learn LLM-based evidentiality, further enhancing performance.

\subsection{Adaptive Compression Ratio Analysis}
\input{table/compression_ratio}
To validate the claim of our adaptive compression capabilities, we analyzed the distribution of compression ratios across datasets. 
The compression ratio is defined as the number of compressed tokens divided by the number of original tokens. 
Table~\ref{tab:compression_ratio_statistics} summarizes the minimum, maximum, mean, median, and standard deviation of compression ratios for the NQ and TQA datasets.

The results highlight differences between datasets, with higher mean and median compression ratios observed for NQ. 
This reflects complexity of dataset, requiring the extraction of answers from lengthy Wikipedia documents through reasoning and comprehensive understanding. 
In contrast, TQA involves documents with explicitly positioned answers, making the task primarily about filtering irrelevant information. 
Consequently, ECoRAG retrieves more evidence for NQ to address its higher information needs, demonstrating its ability to adjust compression ratios adaptively based on dataset complexity and information requirements.

\subsection{Further Analysis on Efficiency}
\input{table/efficiency_comparison}
ECoRAG has demonstrated efficiency over traditional RAG, as shown in Table~\ref{tab:gpt_result} and~\ref{tab:speed_test}, but further analysis is required to verify its resource and latency efficiency.
To compare resource usage, we refer to Table~\ref{tab:efficiency_comparison}.
While traditional RAG requires at least 8B VRAM in our experiments, ECoRAG only adds additional 880M VRAM.
Furthermore, since the compressor and evaluator can operate sequentially as well as simultaneously with the reader, ECoRAG remains feasible in traditional RAG environments.

\input{table/worst_case_latency}
In terms of latency, Table~\ref{tab:speed_test} shows that ECoRAG is more efficient than traditional RAG, but additional verification is needed across different cases. 
The additional modules—compressor and evaluator—may seem to increase system complexity. 
However, traditional RAG must process the entire long context, while ECoRAG reduces latency by 7.32h, as shown in Table~\ref{tab:speed_test}.
Table~\ref{tab:efficiency_comparison} shows that ECoRAG requires little time for compression, reducing the risk of bottleneck as the preceding modules process efficiently.
In the worst case, ECoRAG evaluates compression multiple times, leading to longer latency than in the best case.
However, even in the worst case, Table~\ref{tab:worst_case_latency} demonstrates that ECoRAG is still faster than traditional RAG.

\subsection{Case study of evidentiality-guided compression}
\input{table/case_study}
Table \ref{tab:case_study} illustrates an example of evidentiality-guided compression. 
For the given question, \textit{who dies at the end of Den of Thieves?} with the correct answer \textit{Merrimen}, the initial document set before compression includes the correct answer.
But it also contains irrelevant information, which misleads the LLM into generating the wrong answer, \textit{Donnie.}
After compression, irrelevant content containing the word \textit{Donnie} is effectively suppressed, leaving only the evidential (highlighted) sentences.

\newpage
\subsection{Generalizability across Readers}
\label{sec:appendix_gpt_results}
\input{table/main_results}
\input{table/gemma_results}
\input{table/llama_results}
To evaluate the generalizability of our compression framework, we conducted experiments using Flan-UL2~\cite{tay2023ul} (20B), Llama3~\cite{dubey2024llama} (8B), and Gemma2~\cite{team2024gemma} (9B) as the reader LLMs. 
These models were chosen to investigate how our method performs across diverse architectures and parameter sizes.

Flan-UL2 was selected because RECOMP also utilizes it, as we intend to directly compare with it.
Furthermore, additional experiments were conducted with Llama3 and Gemma2 to extend the evaluation.
Since Llama3 has large context length, it can conduct `standard RAG' experiment, unlike Flan-UL2 and Gemma2.

Results show that our evidentiality-guided compression method consistently outperforms other compression baselines on all three models. 
Specifically, with Flan-UL2 in Table~\ref{tab:main_result}, which was used to define evidentiality during training, the model demonstrated a clear improvement across all metrics. 
Similarly, as shown in Table~\ref{tab:gemma_result}. Gemma2, despite being trained without its own evidentiality mining, also showed improved performance with our compression method, further validating its effectiveness.

In the case of Llama3, as presented in Table~\ref{tab:llama_result}, our compression approach outperformed other baselines, including naive prepend. However, in certain instances, it was outperformed by the `closed book' approach.
This suggests that parametric knowledge embedded within the reader LLM can occasionally align well with specific datasets, leading to variations in performance across models.

Nonetheless, our framework ECoRAG is model-agnostic, as we have excluded the influence of the parametric knowledge of the reader LLM in mining evidentiality labels.
These results emphasize that our compression method consistently outperforms other compression approaches, further validating its effectiveness across diverse models and configurations.

\subsection{Generalizability across Retrievers}
\label{sec:a_9}
\input{table/generalizability_retriever}
To verify that our compression approach generalizes beyond DPR, we conducted additional experiments using another retriever.
Our initial choice of DPR was intentional, in order to demonstrate the robustness of our compression approach even under challenging conditions where a weaker retriever could introduce significant noise.
In Table~\ref{tab:generalizability_retriever}, we then evaluated our method with Contriever~\cite{izacard2022unsupervised}, a stronger dense retriever. 
The results show an even larger performance gain when paired with Contriever than with DPR, indicating that ECoRAG synergizes especially well with higher-quality retrieval.

\input{table/various_doc_size}
To compare against a simple baseline of retrieving fewer documents, we evaluated ECoRAG against varying retrieval sizes.
We initially chose 100 documents to align with standard practice in prior work, such as Fusion-in-Decoder~\cite{izacard2020leveraging}.
In Table~\ref{tab:various_doc_size}, we then conducted experiments on varying numbers of retrieved documents ($k$ = 5, 10, and 20) for both DPR and Contriever, measuring token count and EM without compression.
Our results show that reducing $k$ does not improve accuracy as effectively as adaptive compression via ECoRAG, emphasizing the benefit of our evidentiality-guided approach.

\subsection{Evaluation in Multi-hop QA}
\label{sec:a_10}
\input{table/multi-hop_results}
To assess the effectiveness of ECoRAG in multi-hop QA tasks requiring multiple evidence sources, we conducted experiments in Table~\ref{tab:multi-hop_results}.
ECoRAG classifies evidentiality into three categories and defines weak evidence that supports the correct answer without directly generating the answer.
This enables ECoRAG to perform effectively in tasks requiring partial evidence, such as multi-hop QA.
Furthermore, according to CompAct, adaptively adjusting evidence can collect the partial evidence needed for multi-hop QA, ECoRAG achieves through Evidentiality Reflection.

Table~\ref{tab:multi-hop_results} shows that ECoRAG outperformed both non-compressed and other compression baselines in HotpotQA~\cite{yang-etal-2018-hotpotqa}. 
CompAct and other baselines did not outperform the ``standard RAG'' approach, which uses all 100 documents without compression.
In contrast, ECoRAG improved performance by removing distractors and keeping necessary evidence. 
These results show that ECoRAG is effective for complex scenarios such as multi-hop QA.

\input{table/longbench_results}
To evaluate ECoRAG in scenarios where the challenge lies not only in the number but also in the length of retrieved documents, we applied our method to the LongBench~\cite{bai2024longbench} benchmark. 
LongBench is a long-context understanding benchmark covering tasks such as HotpotQA\footnote{LongBench does not include the full original datasets, so our result in HotpotQA results may differ from those reported in Table~\ref{tab:multi-hop_results}.} and MuSiQue~\cite{trivedi2022musique}.
In Table~\ref{tab:longbench_results}, we compared standard RAG, RECOMP, and ECoRAG (using Llama3-8B) across these tasks within LongBench. 
Consistent with our multi-hop results, ECoRAG outperformed both non-compressed and compression baselines in this long-document setting, further demonstrating its robustness and effectiveness.

%% file: table/contribution.tex
\begin{table*}[t]
    \centering
    \scalebox{1.0}{
    \begin{tabular}{lcccc}
        \noalign{\hrule height 1pt}
        \textbf{Methods} & \begin{tabular}[c]{@{}c@{}}\textbf{Question} \\ \textbf{-aware}\end{tabular} & \begin{tabular}[c]{@{}c@{}}\textbf{Adaptive} \\ \textbf{Compression}\end{tabular} & \begin{tabular}[c]{@{}c@{}}\textbf{Evidentiality} \\ \textbf{-guided}\end{tabular} & \begin{tabular}[c]{@{}c@{}}\textbf{Extractive} \\ \textbf{Compression}\end{tabular} \\
        \hline
        LLMLingua, LLMLingua-2  & \xmark & \xmark & \xmark & \cmark \\ 
        LongLLMLingua  & \cmark & \xmark & \xmark & \cmark \\ 
        RECOMP (extractive) & \cmark & \xmark & \xmark & \cmark \\ 
        RECOMP (abstractive)  & \cmark & \xmark & \xmark & \xmark \\ 
        CompAct  & \cmark & \cmark & \xmark & \xmark \\ 
        ECoRAG (ours)  & \cmark & \cmark & \cmark & \cmark \\
        \noalign{\hrule height 1pt}
    \end{tabular}}
    \caption{The table compares different methods based on their key characteristics. Our approach, ECoRAG, integrates all these features for fast and effective large-scale document compression.}
    \label{tab:comparison}
\end{table*}


%% file: figure/evaluator_metrics_nq.tex
\begin{figure}[h]
{
\centering
    \includegraphics[width=0.8\linewidth]{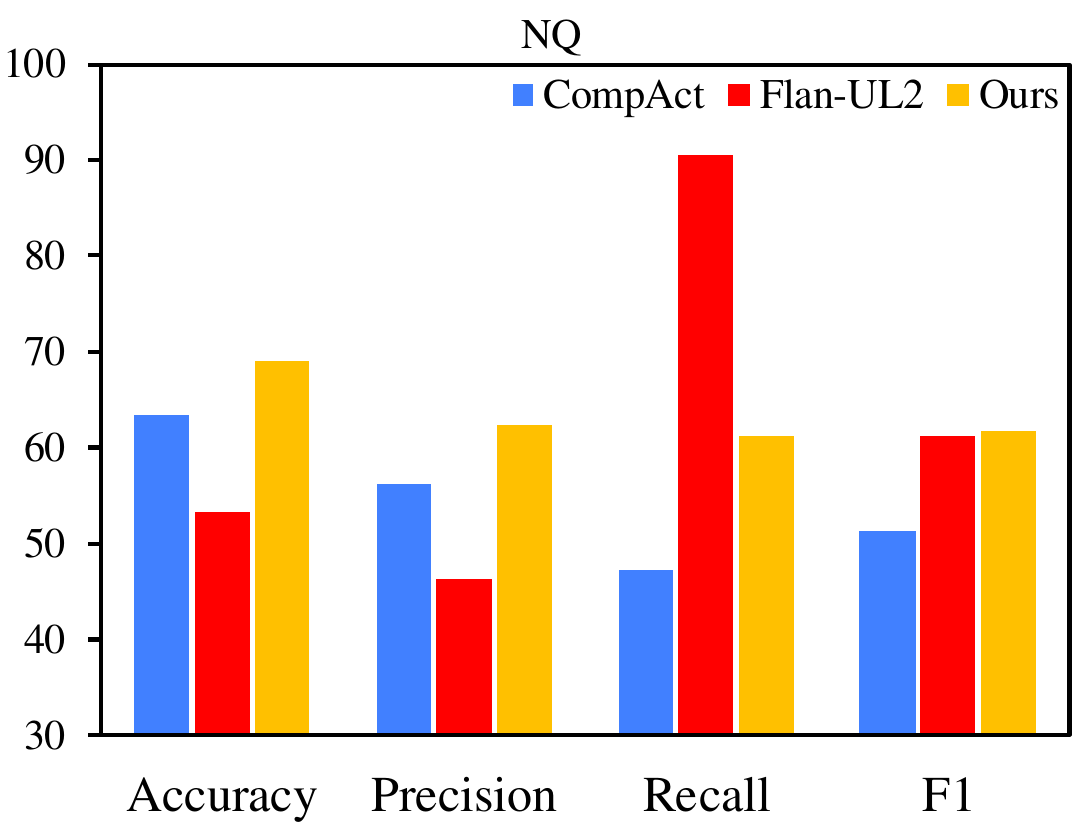}
    \caption{
    Evidentiality evaluation metrics using different evaluator, including ours, measured on the NQ.
    }
    \label{fig:evaluator_metrics_nq}
}
\end{figure}

%% file: table/1000_doc.tex
\begin{table}[]
\centering
\scalebox{0.7}{
\begin{tabular}{lccc}
\noalign{\hrule height 1pt}
\textbf{Methods} & \textbf{\#tokens $\downarrow$} & \textbf{EM} & \textbf{F1} \\
\hline
\textbf{\textit{RAG without compression}} & & & \\
~~ closed-book & 0 & 21.33 & 28.71 \\
~~ standard RAG (1000 documents) & 127,880 & 0.44 & 0.63 \\
\hline
\textbf{\textit{RAG with 1000 documents compressed}} & & & \\
~~ RECOMP (extractive) & 661 & 31.39 & 42.29 \\
~~ ECoRAG (ours) & 659 & \textbf{35.51} & \textbf{48.63} \\
\noalign{\hrule height 1pt}
\end{tabular}
}
\caption{Experimental results on the NQ test dataset using GPT-4o-mini, comparing performance with and without compression for 1000 retrieved documents~\cite{karpukhin-etal-2020-dense}.}
\label{tab:1000_doc}
\end{table}

%% file: table/bge-m3.tex
\begin{table*}[]
\centering
\scalebox{0.9}{
\begin{tabular}{lccccccccc}
\noalign{\hrule height 1pt}
\multirow{2}{*}{\textbf{Methods}} & \multicolumn{3}{c}{\textbf{NQ}} & \multicolumn{3}{c}{\textbf{TQA}} & \multicolumn{3}{c}{\textbf{WQ}} \\
 & \textbf{\#tokens} & \textbf{EM} & \textbf{F1} & \textbf{\#tokens} & \textbf{EM} & \textbf{F1} & \textbf{\#tokens} & \textbf{EM} & \textbf{F1} \\
\hline
BGE-M3 (top 10)          & 330 & 33.02 & 45.47 & 370 & 64.12 & 74.34 & 322 & 20.77 & 38.27 \\
BGE-M3 (top 20)          & 670 & 33.99 & 46.82 & 746 & 65.15 & 75.14 & 645 & 20.77 & 38.00 \\
BGE-reranker (top 10)    & 436 & 34.16 & 47.73 &  –  &   –   &   –   &  –  &   –   &   –   \\
BGE-reranker (top 20)    & 838 & 34.82 & 47.95 &  –  &   –   &   –   &  –  &   –   &   –   \\
ECoRAG (ours)            & 632 & \textbf{36.48} & \textbf{49.81} & 441 & \textbf{65.34} & \textbf{75.37} & 560 & \textbf{30.17} & \textbf{46.13} \\
\noalign{\hrule height 1pt}
\end{tabular}
}
\caption{Performance on NQ, TQA and WQ using GPT-4o-mini, comparing dense retriever BGE-M3, BGE-reranker (results shown only for NQ), and our ECoRAG.}
\label{tab:bge}
\end{table*}

%% file: table/compression_ratio.tex
\begin{table*}[]
\centering
\scalebox{0.85}{
\begin{tabular}{lccccc}
\noalign{\hrule height 1pt}
\textbf{Dataset} & 
\begin{tabular}[c]{@{}c@{}}\textbf{Min Compression} \\ \textbf{Ratio}\end{tabular} & 
\begin{tabular}[c]{@{}c@{}}\textbf{Max Compression} \\ \textbf{Ratio}\end{tabular} & 
\begin{tabular}[c]{@{}c@{}}\textbf{Mean Compression} \\ \textbf{Ratio}\end{tabular} & 
\begin{tabular}[c]{@{}c@{}}\textbf{Median Compression} \\ \textbf{Ratio}\end{tabular} & 
\begin{tabular}[c]{@{}c@{}}\textbf{Standard} \\ \textbf{Deviation}\end{tabular} \\
\hline
NQ & 0.0036 & 1 & 0.0401 & 0.0446 & 0.0247 \\
TQA & 0.0034 & 1 & 0.0267 & 0.0161 & 0.0221 \\
\noalign{\hrule height 1pt}
\end{tabular}
}
\caption{Compression ratio statistics for NQ and TQA datasets.}
\label{tab:compression_ratio_statistics}
\end{table*}

%% file: table/efficiency_comparison.tex
\begin{table}[t]
  \centering
  \resizebox{1.0\linewidth}{!}{
  \begin{tabular}{lccc}
  \noalign{\hrule height 1pt}
  \textbf{} & \textbf{Compressor} & \textbf{Evaluator} & \textbf{Reader} \\
  \hline
  \textbf{VRAM usage} & 110M & 770M & $\geq$8B \\
  \textbf{Latency} & 0.70h & 0.03h & 4.23h \\
  \noalign{\hrule height 1pt}
  \end{tabular}
  }
  \caption{VRAM usage and latency for each component in ECoRAG on the NQ test set.}
  \label{tab:efficiency_comparison}
\end{table}

%% file: table/worst_case_latency.tex
\begin{table}[t]
    \centering
    \resizebox{1.0\linewidth}{!}{
    \begin{tabular}{lcccc}
        \noalign{\hrule height 1pt}
        \textbf{Methods} 
        & \begin{tabular}[c]{@{}c@{}}\textbf{Compression} \\ \textbf{Time}\end{tabular} 
        & \begin{tabular}[c]{@{}c@{}}\textbf{Inference} \\ \textbf{Time}\end{tabular} 
        & \begin{tabular}[c]{@{}c@{}}\textbf{Total} \\ \textbf{Time}\end{tabular} 
        & \begin{tabular}[c]{@{}c@{}}\textbf{Throughput} \\ \textbf{(example/sec)}\end{tabular} \\
        \hline
        standard RAG & - & 8.55h & 8.55h & 0.08  \\
        ECoRAG (ours) & 0.51h & 2.94h & 3.45h & 0.20  \\
        \noalign{\hrule height 1pt}
    \end{tabular}}
    \caption{Inference time and compression time for NQ test under worst case scenarios.}
    \label{tab:worst_case_latency}
\end{table}

%% file: table/case_study.tex
\begin{table*}[ht!]
\scriptsize
\begin{tabular}{p{1.2cm}p{11.5cm}p{1.8cm}}
\toprule
\multicolumn{2}{l}{\textbf{Question}} &  \textbf{Gold answers} \\
\multicolumn{2}{l}{\textit{who dies at the end of den of thieves}} &  \textcolor{blue}{\textbf{Merrimen}} \\
\toprule
\textbf{Type} & \textbf{In-context documents} & \textbf{Prediction} \\ 
\midrule
\textbf{None} &  & \textcolor{red}{\textbf{Donnie}}	 \\ \hline
\textbf{retrieved documents} & 
\small Den of Thieves (film) Nick, forcing Nick to shoot him. 
\hl{As $\textcolor{blue}{\textbf{Merrimen}}$ lies on the ground dying, Nick kneels and consoles him.}
When Nick inspects \textcolor{blue}{\textbf{Merrimen}}'s SUV, he only finds bags with shredded paper; he also finds that \textcolor{red}{\textbf{Donnie}} has escaped custody. Nick later goes to \textcolor{red}{\textbf{Donnie}}'s bar and sees pictures of him with some of the crew members from the heist. It is revealed \textcolor{red}{\textbf{Donnie}} masterminded the heist to keep all of the stolen cash for himself in a second garbage truck. After the passage of some time, \textcolor{red}{\textbf{Donnie}} is working in a London bar, planning a new heist. The film was in Den of Thieves (film) is currently in development. In Los Angeles, a team of robbers led by Ray \textcolor{blue}{\textbf{Merrimen}} make a violent armed attack and hijack an armored truck. Police officers arrive on the scene and engage in a shootout with the robbers. Eventually, \textcolor{blue}{\textbf{Merrimen}} and his crew escape with the empty armored truck. In the morning, Detective Nick O'Brien investigates the crime scene, having been monitoring \textcolor{blue}{\textbf{Merrimen}} and his crew for a while. Suspecting a local bartender named \textcolor{red}{\textbf{Donnie}} for involvement, Nick finds him at the bar and kidnaps him for interrogation. \textcolor{red}{\textbf{Donnie}} reveals \textcolor{blue}{\textbf{Merrimen}} is planning to rob the Federal Reserve on Den of Thieves (film) garbage truck that removes shredded bills. Nick's team catches up to \textcolor{red}{\textbf{Donnie}} and seizes him, beating him until he tells them where \textcolor{blue}{\textbf{Merrimen}} is going. \textcolor{blue}{\textbf{Merrimen}}, Bosco, and Levi try to make their escape with the money bags from the waste truck but hit a traffic jam and are blocked. Nick's team spots them and attempt to shoot them as the robbers try to escape. A shootout occurs initiated by \textcolor{blue}{\textbf{Merrimen}}, killing one of Nick's men. Levi and Bosco are eventually shot dead, but \textcolor{blue}{\textbf{Merrimen}} gets away. Nick chases and shoots \textcolor{blue}{\textbf{Merrimen}}, wounding him. \textcolor{blue}{\textbf{Merrimen}} raises an empty gun to Den of Thieves (film) is currently in development. In Los Angeles, a team of robbers led by Ray \textcolor{blue}{\textbf{Merrimen}} make a violent armed attack and hijack an armored truck. Police officers arrive on the scene and engage in a shootout with the robbers. Eventually, \textcolor{blue}{\textbf{Merrimen}} and his crew escape with the empty armored truck. In the morning, Detective Nick O'Brien investigates the crime scene, having been monitoring \textcolor{blue}{\textbf{Merrimen}} and his crew for a while. Suspecting a local bartender named \textcolor{red}{\textbf{Donnie}} for involvement, Nick finds him at the bar and kidnaps him for interrogation. \textcolor{red}{\textbf{Donnie}} reveals \textcolor{blue}{\textbf{Merrimen}} is planning to rob the Federal Reserve on Den of Thieves (film) Friday of that week by covertly removing about \$30 million in old bills which are scheduled to be shredded after their serial numbers are deleted from computer records. At their hideout, \textcolor{blue}{\textbf{Merrimen}} has one of his crew, Levi, roughly interrogate \textcolor{red}{\textbf{Donnie}} to ensure he didn't disclose anything about the plan. Meanwhile, Nick goes to a strip club and finds \textcolor{blue}{\textbf{Merrimen}}'s stripper girlfriend, hiring her for the night to find out where the heist is going to happen. The next morning, Nick makes an effort to see his daughter at her school. As the day of the heist comes, \textcolor{blue}{\textbf{Merrimen}} and 
  & \textcolor{red}{\textbf{Donnie}} \\ 
\midrule 
\textbf{Compression} & \small Den of Thieves (film) 
\hl{As $\textcolor{blue}{\textbf{Merrimen}}$ lies on the ground dying, Nick kneels and consoles him.} 
Den of Thieves (film) Eventually, \textcolor{blue}{\textbf{Merrimen}} and his crew escape with the empty armored truck. Den of Thieves (film) \textcolor{blue}{\textbf{Merrimen}}, Bosco, and Levi try to make their escape with the money bags from the waste truck but hit a traffic jam and are blocked. Den of Thieves (film) In the morning, Detective Nick O'Brien investigates the crime scene, having been monitoring \textcolor{blue}{\textbf{Merrimen}} and his crew for a while. Den of Thieves (film) Meanwhile, Nick goes to a strip club and finds \textcolor{blue}{\textbf{Merrimen}}'s stripper girlfriend, hiring her for the night to find out where the heist is going to happen.  & \textcolor{blue}{\textbf{Merrimen}} \\
\bottomrule
\end{tabular} \vspace{-0.3em}
\caption{Case study of how the compression of the retrieved documents helps the model to identify the correct answer from NQ test set.
The \hl{highlighted} part is the evidential sentence that directly gives useful information for generating the correct answer \textcolor{blue}{\textbf{Merrimen}}, rather than the incorrect answer \textcolor{red}{\textbf{Donnie}}.}
\label{tab:case_study}
\end{table*}

%% file: table/main_results.tex
\begin{table*}[]
    \centering
    \scalebox{0.78}{
    \begin{tabular}{lcccccccccc}
    \noalign{\hrule height 1pt}
    \multirow{2}{*}{\textbf{Methods}} & \multicolumn{3}{c}{\textbf{NQ}} & \multicolumn{3}{c}{\textbf{TQA}} & \multicolumn{3}{c}{\textbf{WQ}} \\
                             &\textbf{\#tokens $\downarrow$}  & \textbf{EM}     & \textbf{F1} & \textbf{\#tokens $\downarrow$}   & \textbf{EM}   & \textbf{F1} & \textbf{\#tokens $\downarrow$}   & \textbf{EM}   & \textbf{F1} \\ \hline
    \textbf{\textit{RAG without compression}} & \\
    ~~ closed-book       & 0        & 21.33  &   28.71 &         0 &     46.48 & 52.47 & 0 & 32.97 & 42.33 \\
    ~~ standard RAG (100 documents)    &    15456 & -  &   - &     15943 & - & - & 15135 & - & - \\ \hline
    \textbf{\textit{RAG with 100 documents compressed}} & \\
    ~~ LLMLingua               &    725  &   19.17 &   25.48 &       726 & 42.97 & 48.93 & 868 & 31.10 & 40.87 \\
    ~~ LLMLingua-2            &     1475 &   24.63 &   32.19 &      1518 & 53.07 & 59.42 & 1580 & 30.61 & 41.76 \\
    ~~ LongLLMLingua         &    1516 &   38.03 &   46.94 &      1570 & 65.79 & 73.88 & 1629 & 32.78 & 45.27 \\
    ~~ RECOMP (extractive)     &      727 &   38.06 &   46.18 &       750 &  62.49 & 69.68 & 857 & 31.25 & 43.18 \\
    ~~ RECOMP (abstractive)     &      \textbf{16} &  22.22 &   29.56 &  \textbf{30} & 43.50 & 49.88 & \textbf{157} & 38.15 & 38.56  \\
    ~~ CompAct   &         252 &   42.16 &   51.05 &         253 &   64.37   &   72.25  & 218 & 33.07 & 44.45 \\
    ~~ ECoRAG (ours)             &         693 & \textbf{44.38}  &  \textbf{53.56}  &          501 &   \textbf{66.45}   &   \textbf{74.02}  & 671 & \textbf{33.71} & \textbf{46.08} \\
    \noalign{\hrule height 1pt}
    \end{tabular}
    }
    \caption{Comparison of compression methods on NQ, TQA, and WQ using Flan-UL2~\cite{tay2023ul} with 100 retrieved documents~\cite{karpukhin-etal-2020-dense}.}
    \label{tab:main_result}
\end{table*}

%% file: table/gemma_results.tex
\begin{table*}[]
    \centering
    \scalebox{0.78}{
    \begin{tabular}{lcccccccccc}
    \noalign{\hrule height 1pt}
    \multirow{2}{*}{\textbf{Methods}} & \multicolumn{3}{c}{\textbf{NQ}} & \multicolumn{3}{c}{\textbf{TQA}} & \multicolumn{3}{c}{\textbf{WQ}} \\
                             &\textbf{\#tokens $\downarrow$}  & \textbf{EM}     & \textbf{F1} & \textbf{\#tokens $\downarrow$}   & \textbf{EM}   & \textbf{F1} & \textbf{\#tokens $\downarrow$}   & \textbf{EM}   & \textbf{F1} \\ \hline
    \textbf{\textit{RAG without compression}} & \\
    ~~ closed-book       & 0        & 27.84  &   38.35 &         0 &     57.11 & 66.39 & 0 & 26.77 & 43.24 \\
    ~~ standard RAG (100 documents)    &    14260 & -  &   - &     - & - & - & 14075 & - & - \\ \hline
    \textbf{\textit{RAG with 100 documents compressed}} & \\
    ~~ LLMLingua               &    643  &   26.90 &   37.90 &       638 & 60.71 & 68.09 & 649 & 25.04 & 42.08 \\
    ~~ LLMLingua-2            &     1403 &   28.56 &   38.95 &      1393 & 59.95 & 67.84 & 1401 & 24.36 & 40.52 \\
    ~~ LongLLMLingua         &    1411 &   37.67 &   49.40 &      1436 & 63.17 & 70.28 & 1399 & 27.02 & 44.23 \\
    ~~ RECOMP (extractive)     &      165 &   37.65 &   48.24 &       687 &  63.19 & 70.38 & 680 & 26.03 & 42.22 \\
    ~~ RECOMP (abstractive)     &      \textbf{17} &  27.98 &   38.00 &  \textbf{28} & 58.78 & 65.74 & \textbf{21} & 25.20 & 41.60 \\
    ~~ CompAct   &         111 &   38.67 &   49.87 &         100 &   65.88   &   73.29  & 78 & 26.67 & 43.04 \\
    ~~ ECoRAG (ours)             &         684 & \textbf{39.20}  &  \textbf{50.24}  &          448 &   \textbf{66.32}   &   \textbf{74.25}  & 504 & \textbf{27.41} & \textbf{44.00} \\
    \noalign{\hrule height 1pt}
    \end{tabular}
    }
    \caption{Comparison of compression methods on NQ, TQA, and WQ using Gemma2~\cite{team2024gemma} with 100 retrieved documents~\cite{karpukhin-etal-2020-dense}.}
    \label{tab:gemma_result}
\end{table*}

%% file: table/llama_results.tex
\begin{table*}[]
    \centering
    \scalebox{0.78}{
    \begin{tabular}{lcccccccccc}
    \noalign{\hrule height 1pt}
    \multirow{2}{*}{\textbf{Methods}} & \multicolumn{3}{c}{\textbf{NQ}} & \multicolumn{3}{c}{\textbf{TQA}} & \multicolumn{3}{c}{\textbf{WQ}} \\
                             &\textbf{\#tokens $\downarrow$}  & \textbf{EM}     & \textbf{F1} & \textbf{\#tokens $\downarrow$}   & \textbf{EM}   & \textbf{F1} & \textbf{\#tokens $\downarrow$}   & \textbf{EM}   & \textbf{F1} \\ \hline
    \textbf{\textit{RAG without compression}} & \\
    ~~ closed-book       & 0        & 22.16  &   32.36 &         0 &     \textbf{60.89} & 67.80 & 0 & \textbf{21.79} & \textbf{35.81} \\
    ~~ standard RAG (100 documents)    &    14263 & 0.27  &   0.97 &     14574 & 0.24 & 2.70 & 14147 & 0.25 & 4.48 \\ \hline
    \textbf{\textit{RAG with 100 documents compressed}} & \\
    ~~ LLMLingua               &    641  &   15.20 &   22.31 &       636 & 52.11 & 59.23 & 646 & 17.62 & 30.92 \\
    ~~ LLMLingua-2            &     1346 &   3.91 &   7.19 &      1366 & 48.08 & 55.91 & 1337 & 4.28 & 11.44 \\
    ~~ LongLLMLingua         &    1388 &   20.30 &   28.85 &      1423 & 58.34 & 68.49 & 1372 & 18.70 & 32.12 \\
    ~~ RECOMP (extractive)     &      160 &   22.33 &   31.12 &       683 &  36.69 & 44.08 & 667 & 16.19 & 27.80 \\
    ~~ RECOMP (abstractive)     &      \textbf{16} &  18.75 &   27.85 &  \textbf{27} & 42.73 & 50.94 & \textbf{21} & 18.80 & 33.25 \\
    ~~ CompAct   &         107 &   28.01 &   38.52 &         99 &   56.01   &   64.69  & 76 & 21.41 & 35.21 \\
    ~~ ECoRAG (ours)             &         519 & \textbf{30.22}  &  \textbf{42.55}  &          445 &   59.25   &   \textbf{69.32}  & 588 & 21.60 & 35.43 \\
    \noalign{\hrule height 1pt}
    \end{tabular}
    }
    \caption{Comparison of compression methods on NQ, TQA, and WQ using Llama3~\cite{dubey2024llama} with 100 retrieved documents~\cite{karpukhin-etal-2020-dense}.}
    \label{tab:llama_result}
\end{table*}

%% file: table/generalizability_retriever.tex
\begin{table}[]
\centering
\resizebox{1.0\linewidth}{!}{
\begin{tabular}{lccc}
\noalign{\hrule height 1pt}
\textbf{Methods} & \textbf{\#tokens $\downarrow$} & \textbf{EM} & \textbf{F1} \\
\hline
\textbf{\textit{RAG without compression}} & & & \\
~~ closed-book & 0     & 31.88 & 44.10 \\
~~ standard RAG (100 documents) & 13,847 & 37.11 & 50.82 \\
\hline
\textbf{\textit{RAG with 100 documents compressed}} & & & \\
~~ LLMLingua                   & 645   & 25.79 & 37.56 \\
~~ LLMLingua-2                 & 1,319 & 29.95 & 44.40 \\
~~ LongLLMLingua               & 1,364 & 33.44 & 46.20 \\
~~ RECOMP (extractive)         & 659   & 33.21 & 45.98 \\
~~ RECOMP (abstractive)       & \textbf{16}    & 30.45 & 43.01 \\
~~ CompAct                     & 75    & 37.69 & 51.65 \\
~~ ECoRAG (ours)               & 641   & \textbf{41.43} & \textbf{54.02} \\
\noalign{\hrule height 1pt}
\end{tabular}
}
\caption{Experimental results on the NQ dataset using GPT-4o-mini, comparing performance with and without compression for documents retrieved by Contriever.}
\label{tab:generalizability_retriever}
\end{table}

%% file: table/various_doc_size.tex
\begin{table}[]
\centering
\resizebox{1.0\linewidth}{!}{
\begin{tabular}{lcccc}
\noalign{\hrule height 1pt}
\textbf{Methods} & 
\begin{tabular}[c]{@{}c@{}}\textbf{\#tokens}\\\textbf{(DPR)}\end{tabular} & 
\begin{tabular}[c]{@{}c@{}}\textbf{EM}\\\textbf{(DPR)}\end{tabular} & 
\begin{tabular}[c]{@{}c@{}}\textbf{\#tokens}\\\textbf{(Contriever)}\end{tabular} & 
\begin{tabular}[c]{@{}c@{}}\textbf{EM}\\\textbf{(Contriever)}\end{tabular} \\
\hline
\textbf{\textit{Reduced retrieval size}} & & & & \\
~~ \# docs (k) = 5   &  693  & 35.53 &  690  & 33.48 \\
~~ \# docs (k) = 10  & 1,386 & 35.95 & 1,381 & 34.76 \\
~~ \# docs (k) = 20  & 2,774 & 36.33 & 2,762 & 36.47 \\
\hline
\textbf{\textit{Adaptive compression}} & & & & \\
~~ ECoRAG (ours) & \textbf{632} & \textbf{36.48} & \textbf{641} & \textbf{41.43} \\
\noalign{\hrule height 1pt}
\end{tabular}
}
\caption{Experimental results on the NQ dataset using GPT-4o-mini, comparing reduced retrieval sizes for DPR and Contriever against adaptive compression via ECoRAG.}
\label{tab:various_doc_size}
\end{table}

%% file: table/multi-hop_results.tex
\begin{table}[]
\centering
\resizebox{1.0\linewidth}{!}{
\begin{tabular}{lccc}
\noalign{\hrule height 1pt}
\textbf{Methods} & \textbf{\#tokens $\downarrow$} & \textbf{EM} & \textbf{F1} \\
\hline
\textbf{\textit{RAG without compression}} & & & \\
~~ closed-book & 0 & 26.19 & 36.71 \\
~~ standard RAG (100 documents) & 14,313 & 34.52 & 44.69 \\
\hline
\textbf{\textit{RAG with 100 documents compressed}} & & & \\
~~ LLMLingua & 636 & 22.57 & 31.54 \\
~~ LLMLingua-2 & 1,330 & 26.66 & 37.00 \\
~~ LongLLMLingua & 1,406 & 27.45 & 38.07 \\
~~ RECOMP (extractive) & 688 & 28.05 & 38.87 \\
~~ RECOMP (abstractive) & \textbf{12} & 24.27 & 33.88 \\
~~ CompAct & 74 & 31.21 & 42.42 \\
~~ ECoRAG (ours) & 647 & \textbf{34.69} & \textbf{45.13} \\
\noalign{\hrule height 1pt}
\end{tabular}
}
\caption{Experimental results on the HotpotQA dataset using GPT-4o-mini, comparing performance with and without compression for 100 documents~\cite{karpukhin-etal-2020-dense}.}
\label{tab:multi-hop_results}
\end{table}

%% file: table/longbench_results.tex
\begin{table}[]
\centering
\resizebox{1.0\linewidth}{!}{
\begin{tabular}{lcc}
\noalign{\hrule height 1pt}
\textbf{Method} & \textbf{HotpotQA} & \textbf{MusiQue} \\
\hline
standard RAG   & 49.76 & 23.91 \\
RECOMP         & 49.12 & 23.08 \\
ECoRAG (ours)  & \textbf{52.29} & \textbf{24.60} \\
\noalign{\hrule height 1pt}
\end{tabular}
}
\caption{Experiments on LongBench multi-hop datasets (HotpotQA and MusiQue) evaluating F1-score performance of standard RAG, RECOMP, and ECoRAG using Llama3-8B.}
\label{tab:longbench_results}
\end{table}

%% file: content/B_details.tex
\section{Experimental Details}
\label{sec:appendix}
\subsection{Implementation Details}
We used 8 Nvidia RTX3090 GPUs to train all models. 
For mining evidentiality labels for all sentences in retrieved documents, we used the NLTK library \footnote{\href{https://www.nltk.org/}{www.nltk.org}} to split DPR \cite{karpukhin-etal-2020-dense} retrieved top-100 documents into sentences.
To reduce costs, we used the open LLM Flan-UL2\footnote{
\href{https://huggingface.co/google/flan-ul2}{google/flan-ul2}}~\cite{tay2023ul}, which was also used in our experiments and RECOMP~\cite{xu2024recomp}, to label evidentiality based on the definition in Section~\ref{sec:3_1_1}.

Our evidentiality compressor was trained from Contriever~\cite{izacard2022unsupervised} checkpoint pretrained on CC-net \cite{wenzek-etal-2020-ccnet} and English Wikipedia \cite{izacard2022unsupervised}..
We trained it using the AdamW optimizer with a batch size of 64 and a learning rate of $5 \cdot 10^{-5}$ for 4 epochs on NQ~\cite{kwiatkowski2019natural} and WQ~\cite{berant2013semantic}, and 2 epochs on TQA~\cite{joshi-etal-2017-triviaqa}.
While training with $\mathcal{L}_{we}$ and $\mathcal{L}_{se}$ losses, we we used 8 positive contexts and 56 negative contexts per batch.
When calculating the $\mathcal{L}_{se}$ loss, we used negative set with weak evidence to distractor ratio of 0.15:0.85, treating weak evidence as hard negative. 
We set the temperature $\tau$ for the contrastive loss to 1.0.

Our evidentiality evaluator was trained from a pretrained Flan-T5-large checkpoint\footnote{\href{https://huggingface.co/google/flan-t5-large}{google/flan-t5-large}} using the AdamW optimizer.
We trained it with a batch size of 40 and a learning rate of $1 \cdot 10^{-5}$ for 4 epochs with all datasets.
We included `<NOT>' sentences with high compressor scores in the training stage to make the evidentiality evaluator distinguish only the genuinely strong evidence `<EVI>' from the seemingly plausible ones.
We constructed the training data for the evaluator with a ratio of 1:3 between `<EVI>' and `<NOT>' sentences.
For adaptive compression, a limit on the number of evidence pieces was necessary to avoid infinite loops, which we set at 20.
We set this limit to 20 to achieve a compression level similar to RECOMP, but it can be increased for tasks that require more evidence.
Additionally, to prevent high latency due to overly frequent evaluations, we incrementally added 4 evidence pieces at a time.
For experiments on the test set, we used GPT-4o-mini\footnote{gpt-4o-mini-2024-07-18}, Flan-UL2, Gemma2\footnote{\href{https://huggingface.co/google/gemma-2-9b-it}{google/gemma-2-9b-it}}, and Llama3\footnote{\href{https://huggingface.co/meta-llama/Meta-Llama-3-8B-Instruct}{meta-llama/Meta-Llama-3-8B-Instruct}}.

\subsection{Selection for Evidentiality Evaluator}
\label{sec:b_2}
\input{table/evaluator_exposure}
We chose Flan-T5-large as the basis for our Evidentiality Evaluator due to its strong instruction-finetuning and robust performance on classification tasks.
T5-large has been widely used in prior RAG research~\cite{han2023monotonic, yan2024corrective} for document-based evaluation.
For our evidentiality scoring task, we employ Flan-T5-large as it demonstrates enhanced instruction-following capabilities that are well-suited for this classification task. 
However, a potential concern arises regarding Flan-T5-large's prior exposure to datasets such as NQ, TQA, and WQ, which might lead to memorization rather than genuine evidentiality learning. As reported in research related to Flan-T5-large~\cite{chung2022scaling, wang2022super}, its exposure was clearly separated into exposed and non-exposed subsets, and our comparison experiments (Section~\ref{sec:5_2}) demonstrate negligible performance differences between these groups. 
When we applied a two-proportion Z test to the results in Table~\ref{tab:evaluator_explosure}, the analysis at the 0.05 significance level confirmed that the observed differences are not statistically significant. 
Therefore, these observation indicates that the model has learned evidentiality principles rather than simply memorizing evidence from prior exposure.

\subsection{Input Prompts for LLM}
\label{sec:prompt}
\input{prompt/prompt_qa}
\input{prompt/prompt_evi}
We report two examples of input prompts for reader LLMs. 
In Figure~\ref{prompt:prompt_qa}, we report the input prompt used for evidentiality mining and test set experiments to answer a given question when provided with the question and the compressed documents.
This prompt was also utilized during the evidentiality mining process, as described in Section \ref{sec:3_1_1}.
Figure~\ref{prompt:prompt_evi} presents the input prompt for mining the ground truth label of compressed documents using Flan-UL2 as the evidentiality evaluator in the experiments detailed in Section \ref{sec:5_2}.

\section{Usage of AI Assistants}
We utilized ChatGPT to improve the clarity and grammatical accuracy of my writing. It provided suggestions for rephrasing sentences and correcting grammatical errors to make the text flow more naturally.

%% file: table/evaluator_exposure.tex
\begin{table}[]
\centering
\resizebox{1.0\linewidth}{!}{
\begin{tabular}{lcccc}
\noalign{\hrule height 1pt}
\textbf{Subset} & \textbf{Accuracy} & \textbf{Precision} & \textbf{Recall} & \textbf{F1-score} \\
\hline
exposed subset       & 68.52 & 81.47 & 64.17 & 72.13 \\
non-exposed subset   & 71.77 & 78.64 & 76.22 & 77.42 \\
\noalign{\hrule height 1pt}
\end{tabular}
}
\caption{Evaluation results of the Flan-T5-based evidentiality evaluator on TQA, comparing performance between the exposed and non-exposed subsets.}
\label{tab:evaluator_explosure}
\end{table}

%% file: prompt/prompt_qa.tex
\begin{figure*}[t]
    \centering
    \begin{tcolorbox}[title=\textbf{Question Answering Prompt}]
        who won a million on deal or no deal \\
        Answer: Tomorrow Rodriguez \\
        
        who is the woman washing the car in cool hand luke \\
        Answer: Joy Harmon \\
        
        who is the actor that plays ragnar on vikings \\
        Answer: Travis Fimmel \\
        
        who said it's better to have loved and lost \\
        Answer: Alfred , Lord Tennyson \\
        
        name the first indian woman to be crowned as miss world \\
        Answer: Reita Faria \\
        
        \texttt{Documents} \\
        \texttt{Question} \\
        Answer: 
    \end{tcolorbox}
    \caption{An input prompt for LLM for question answering, including few-shot examples, input documents, and a question.}
    \label{prompt:prompt_qa}
\end{figure*}

%% file: prompt/prompt_evi.tex
\begin{figure*}[t]
    \centering
    \begin{tcolorbox}[title=\textbf{Evidentiality Evaluation Prompt}]
        You are an expert at determining whether a document provides evidential support for a given question. You will receive a question and a document, and your task is to evaluate whether the document is evidential, partially evidential, or non-evidential in relation to the question.\\
        Assess the support provided by the document using the following scale:\\
        - [Evidential] - The document fully supports the question, providing clear and direct evidence that answers or addresses the query completely.\\
        - [Non-Evidential] - The document does not provide relevant information or evidence related to the question, making it unrelated or insufficient to support the query.\\
        Please provide your assessment and briefly justify your reasoning based on the content of the document in relation to the question.\\
        
        Question: what is the temperature of dry ice in kelvin?\\
        Evidence: At atmospheric pressure, sublimation/deposition occurs at or 194.65 K. The density of dry ice varies, but usually ranges between about.\\
        Score: [Evidential]\\
        
        Question: when did north vietnam unify with the south?\\
        Evidence: The distinctive synthesizer theme was performed by the then-little-known Thomas Dolby, and this song also marked a major departure from their earlier singles because their previous singles were mid to upper tempo rock songs while this song was a softer love song with the energy of a power ballad.\\
        Score: [Non-Evidential]\\
        
        Question: who played all the carly 's on general hospital?\\
        Evidence: Throughout the 2000s, Carly, then Tamara Braun (2001–05) goes on to become one of the\\
        Score: [Non-Evidential]\\
        
        Question: who sang the original blinded by the light?\\
        Evidence: Light of Day (song) "Light of Day", sometimes written as "(Just Around the Corner to the) Light of Day", is a song written by Bruce Springsteen and performed initially by Joan Jett and Michael J.\\
        Score: [Non-Evidential]\\
        
        Question: who was the rfc editor until 1998 just provide the family name?\\
        Evidence: Perhaps his most famous legacy is from RFC 760, which includes a robustness principle often called "Postel's law": "an implementation\\
        Score: [Non-Evidential]\\
        
        Question: \texttt{Question} \\
        Evidence: \texttt{Compressed Documents} \\
        Score: 
    \end{tcolorbox}
    \caption{An input prompt for LLM for evidentiality evaluation, including few-shot examples, compressed documents, and a question.}
    \label{prompt:prompt_evi}
\end{figure*}